\documentclass{article} % For LaTeX2e
\usepackage{iclr2026_conference,times}

\usepackage{makecell}
\usepackage[table,dvipsnames]{xcolor}
\definecolor{myPink}{rgb}{0.980, 0.502, 0.447} 
\definecolor{myOrange}{rgb}{1.000, 0.784, 0.486} 
\colorlet{colorFst}{myPink!40} % first 
\colorlet{colorSnd}{myOrange!45} % second %
\colorlet{colorTrd}{Goldenrod!20} % third
\colorlet{colorLow}{darkgray!60}    % low-light color
\newcommand{\st}{\cellcolor{colorFst}\bf}   % first
\newcommand{\nd}{\cellcolor{colorSnd}}      % second
\newcommand{\trd}{\cellcolor{colorTrd}}      % third
\newcommand{\lo}{\color{colorLow}}          % low-light
\usepackage{amsmath,amsfonts,bm}

\def\eqref#1{equation~\ref{#1}}
\def\1{\bm{1}}

\DeclareMathAlphabet{\mathsfit}{\encodingdefault}{\sfdefault}{m}{sl}
\SetMathAlphabet{\mathsfit}{bold}{\encodingdefault}{\sfdefault}{bx}{n}

\usepackage[accsupp]{axessibility}
\usepackage{arydshln}
\usepackage{bm}
\usepackage{boldline}
\usepackage{enumitem}
\usepackage{float}
\usepackage{multirow}
\usepackage{nicefrac}
\usepackage{pifont}
\usepackage[most]{tcolorbox}
\usepackage{mathtools}
\usepackage{makecell}
\usepackage{url}
\usepackage{booktabs}
\usepackage{colortbl}               % ensure \cellcolor is defined
\usepackage[table]{xcolor}
\usepackage{stfloats}
\usepackage{graphicx}
\usepackage{subfig}
\usepackage{wrapfig}

\definecolor{Urlcolor}{RGB}{251,111,146}
\definecolor{Linkcolor}{RGB}{193,18,31}
\definecolor{CiteColor}{RGB}{32,126,190}
\definecolor{darkgreen}{RGB}{0.0, 0.5, 0.0}
\usepackage[pagebackref,breaklinks,colorlinks,urlcolor=Urlcolor,linkcolor=Linkcolor,citecolor=CiteColor]{hyperref}
\usepackage{cleveref}

\newcommand{\rebuttal}[1]{{{#1}}}

\newcommand{\method}{NOVA3R}

\title{\method: Non-pixel-aligned Visual Transformer for Amodal 3D Reconstruction}

\author{Weirong Chen$^{1,2}$,
\textbf{Chuanxia Zheng$^{3,4}\thanks{Project Lead}$} ,
\textbf{Ganlin Zhang$^{1,2}$},
\textbf{Andrea Vedaldi$^{3}$},
\textbf{Daniel Cremers$^{1,2}$} \\
$^{1}$Technical University of Munich \quad 
$^{2}$Munich Center for Machine Learning \\
$^{3}$University of Oxford \quad 
$^{4}$Nanyang Technological University \\ 
}
\iclrfinalcopy % Uncomment for camera-ready version, but NOT for submission.
\begin{document}

\maketitle

\begin{abstract}
We present NOVA3R, an effective approach for non-pixel-aligned 3D reconstruction from a set of unposed images in a feed-forward manner.
Unlike pixel-aligned methods that tie geometry to per-ray predictions,
our formulation learns a global, view-agnostic scene representation that decouples reconstruction from pixel alignment.
This addresses two key limitations in pixel-aligned 3D:
(1) it recovers both visible and invisible points with a complete scene representation, and (2) it produces physically plausible geometry with fewer duplicated structures in overlapping regions.
To achieve this, we introduce a scene-token mechanism that aggregates information across unposed images and a diffusion-based 3D decoder that reconstructs complete, non-pixel-aligned point clouds.
Extensive experiments on both scene-level and object-level datasets demonstrate that NOVA3R outperforms state-of-the-art methods in terms of reconstruction accuracy and completeness. 
Our project page is available at: \url{https://wrchen530.github.io/nova3r}.
\end{abstract}

\begin{figure*}[h]
    \vspace{-1em}
    \centering
    \includegraphics[width=0.95\linewidth]{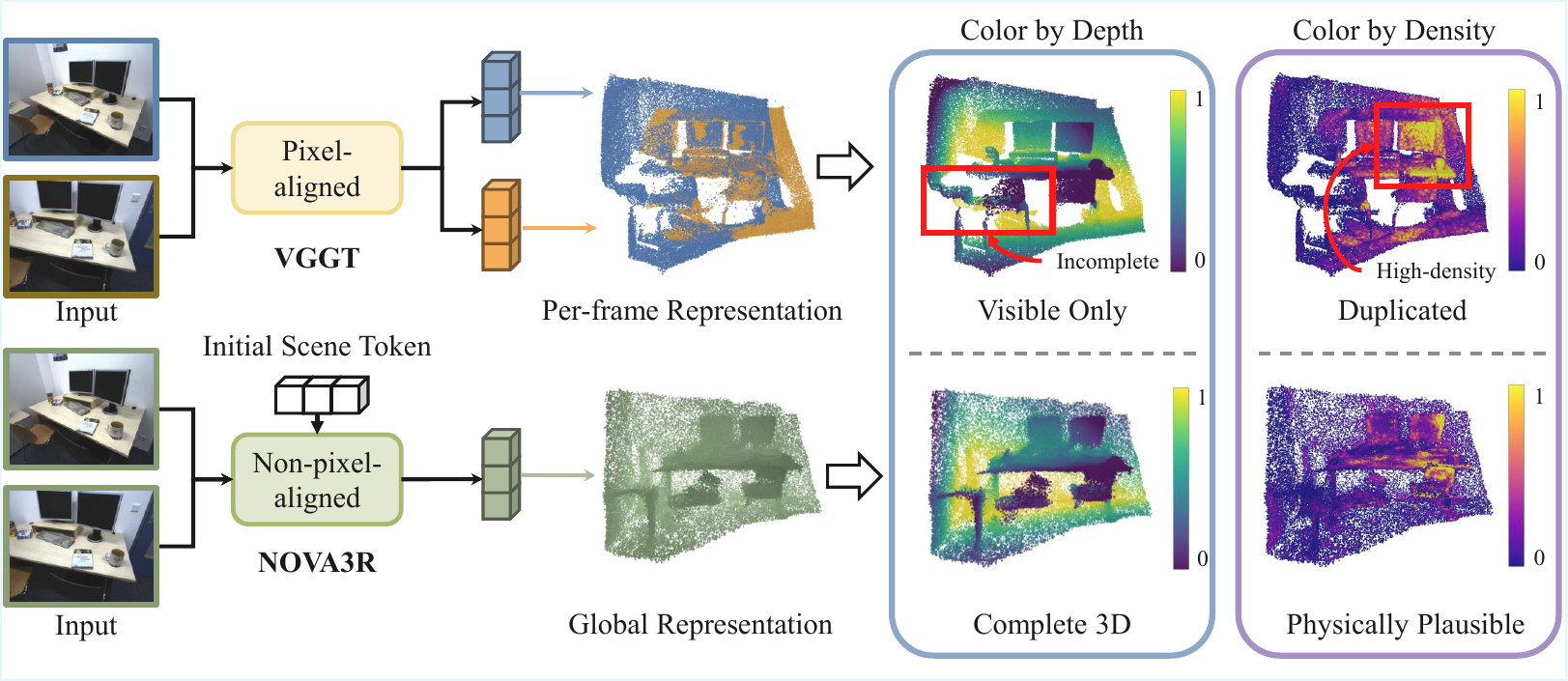}
      \vspace{-1em}
    \caption{\textbf{\method} enables non–pixel-aligned reconstruction by learning a global scene representation from unposed images. Compared to pixel-aligned methods, \method~recovers both visible and occluded regions and produces more physically plausible geometry with fewer duplicated structures. }%
    \label{fig:teaser}
\end{figure*}

\section{Introduction}
\label{sec:intro}

We consider the problem of \emph{non–pixel-aligned} 3D reconstruction from one or more unposed images,
in a feed-forward manner.
This is a challenging task, as the model must infer a global,
view-agnostic representation of the scene without relying on per-ray supervision.
This formulation avoids the limitations of pixel-aligned methods,
which reconstruct only visible surfaces and often produce redundant geometry in overlapping regions.
It therefore enables more complete and physically plausible 3D reconstruction,
capturing both visible and occluded structures in a consistent manner.

Recent work in 3D reconstruction has largely focused on the \emph{pixel-aligned} formulation, where geometry is predicted in the form of depth maps, point maps, or radiance fields tied to the image plane. 
DUSt3R~\citep{wang2024dust3r} pioneers this paradigm of dense,
pixel-aligned 3D reconstruction from unposed image collections,
achieving impressive results in reconstructing the visible regions of a scene.
Building on this,
follow-up works~\citep{tang2025mv,wang2025continuous,Yang_2025_Fast3R,zhang2025FLARE,wang2025vggt}
extend DUSt3R from image pairs to multi-view settings, enabling feed-forward 3D geometry reconstruction from larger image sets.
However, the pixel-aligned formulation remains tied to per-ray prediction,
which restricts reconstruction to visible regions and yields \emph{incomplete} geometry and \emph{overlapping point layers} in areas visible to multiple cameras.

Another line of work explores latent 3D generation, which learns a \emph{global representation} in a compact latent space and decodes it into voxels or meshes~\citep{vahdat2022lion,zhang20233dshape2vecset,zhang2024clay,ren2024xcube,xiang2024structured,TripoSR2024,yang2024hunyuan3d,hunyuan3d22025tencent,li2025triposg}.
While this global formulation can plausibly complete occluded regions beyond the input views,
most approaches remain confined to the \emph{object level}.
They assume canonical space and require high-quality mesh supervision,
which makes these methods struggle with complex, cluttered scenes.
For \emph{scene}-level reconstruction,
some methods~\citep{chen2024mvsplat360,liu2024reconx,gao2024cat3d,szymanowicz2025bolt3d} inpaint unseen regions by synthesizing novel views with pre-trained diffusion models and then post-process to recover geometry. However, such pipelines do not guarantee physically meaningful point clouds.

To overcome these limitations,
we introduce the Non-pixel-aligned Visual Transformer (\method) (see~\Cref{fig:teaser}).
First, we address the challenge of non-pixel-aligned supervision by leveraging a diffusion-based
% 3D latent
3D autoencoder.
It first compresses complete point clouds into compact latent tokens,
and then decodes them back into the original space,
supervised with a flow-matching loss that resolves matching ambiguities in unordered point sets.
Recent works on 3D autoencoders~\citep{zhang20233dshape2vecset,xiang2024structured,yang2024hunyuan3d,li2025triposg} have demonstrated the effectiveness of latent representations,
but they are primarily designed for object reconstruction,
assuming high-quality meshes for supervision.
In contrast, our formulation targets scene-level reconstruction and requires only point clouds derived from meshes or depth maps for supervision,
enabling it to capture priors of complete 3D scenes and produce physically coherent geometry without duplicated points.

Second,
we tackle the problem of mapping unposed images to a global scene representation.
Training such a model directly would require massive amounts of complete scene data and computational resources.
To improve generalization,
our model is built on a pre-trained image encoder from VGGT~\citep{wang2025vggt},
augmenting it with learnable scene tokens that aggregate information from arbitrary numbers of views and map them into the latent space of our point decoder.
This design enables \method~to support both monocular and multi-view reconstruction,
without being restricted to a fixed number of inputs.
Despite being trained on relatively small datasets, our model generalizes well to unseen scenes,
achieving complete and physically plausible reconstructions.

\begin{figure*}[tb!]
    \centering
    \includegraphics[width=0.95\linewidth]{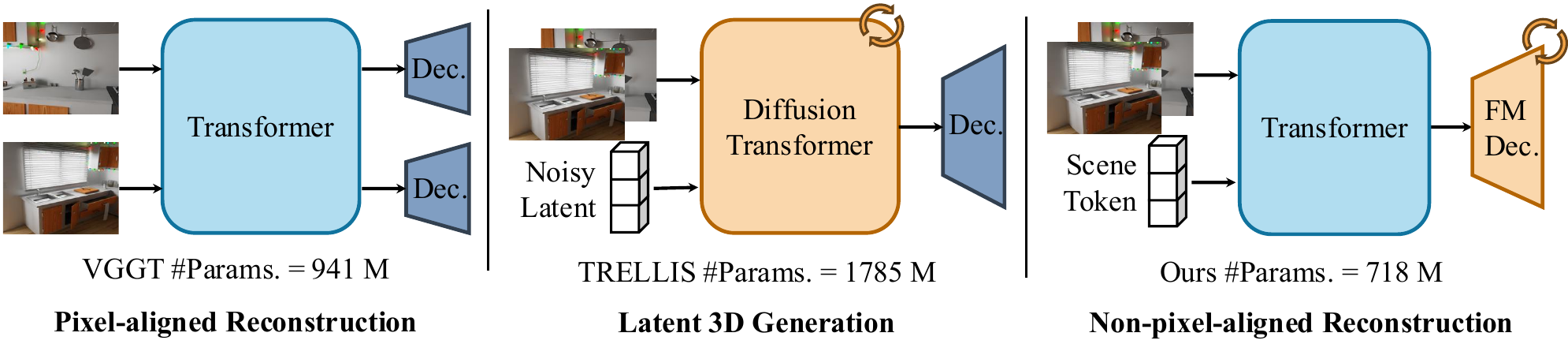}
    \vspace{-10pt}
    \caption{\textbf{Comparison of different reconstruction paradigms.} 
    Our non-pixel-aligned approach combines feed-forward efficiency with a global, view-agnostic scene representation, removing the reliance on pixel-level supervision. \method ~provides a unified solution for various reconstruction tasks, achieving multi-view consistency and geometrically faithful results.}
    \label{fig:motivation}
    \vspace{-10pt}
\end{figure*}

In summary, our main contributions are as follows:
(i) We introduce a unified non-pixel-aligned reconstruction pipeline with minimal assumptions, applicable to both object-level and scene-level complete reconstruction tasks.
(ii) We address key limitations of pixel-aligned methods, which often produce incomplete point clouds, duplicated geometry, and 3D inconsistencies in overlapping regions. By contrast, our non-pixel-aligned formulation naturally yields complete and evenly distributed geometry.
(iii) We integrate a feed-forward transformer architecture with a lightweight flow-matching decoder, effectively bridging the gap between pixel-aligned reconstruction and latent 3D generation, combining feed-forward efficiency with strong 3D modeling capability (see~\Cref{fig:motivation}).

\section{Related Work}
\label{sec:related}

\paragraph{Feed-Forward 3D Reconstruction.}
% Ensure your main .tex file includes the bibliography:
Unlike \emph{per-scene} optimization methods~\citep{mildenhall2020nerf,kerbl20233d} that iteratively refine a 3D representation for each individual scene,
\emph{feed-forward} 3D reconstruction approaches aim to generalize across scenes by predicting 3D geometry directly from a set of input images in a single pass of a neural network.
Early approaches
typically focus on predicting geometric representations,
such as depth maps~\citep{eigen2015predicting}, meshes~\citep{wang2018pixel2mesh}, point clouds~\citep{fan2017point}, or voxel grids~\citep{choy20163d},
and are trained on relatively small-scale datasets~\citep{Silberman:ECCV12,chang2015shapenet}.
As a result, these models struggled to capture fine-grained visual appearance and exhibited limited generalization to unseen scenes.

More recently,
DUSt3R~\citep{wang2024dust3r}
and MASt3R~\citep{leroy2024grounding}
directly regress dense, pixel-aligned point maps from unposed image collections.
These approaches mark a significant step toward generalizable, pose-free 3D reconstruction.
Building on this paradigm,
many recent works~\citep{tang2025mv,wang2025continuous,Yang_2025_Fast3R,zhang2025FLARE,wang2025vggt}
extend it from image pairs to multi-view settings,
enabling feed-forward 3D geometry reconstruction from sets of uncalibrated images.
However, these pixel-aligned methods produce incomplete geometry and duplicated points in overlapping regions.
In contrast,
our approach outputs a unified and \emph{complete} 3D reconstruction that integrates both \emph{visible} and \emph{occluded} regions.

\begin{figure*}[tb!]
    \centering
    \includegraphics[width=\linewidth]{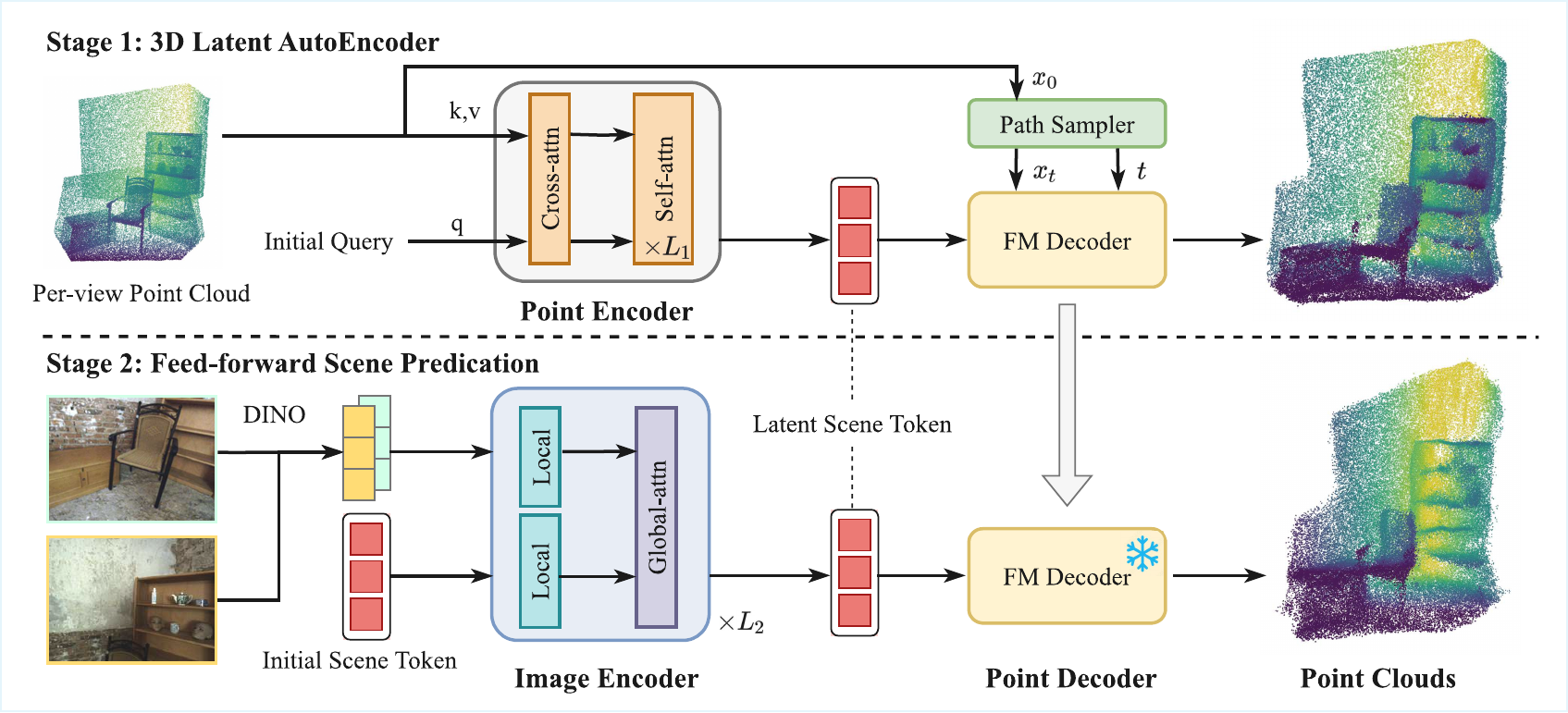}
    \caption{\textbf{Overview of \method.}
    \textbf{Stage 1:} a 3D point autoencoder encodes complete point clouds into latent scene tokens and decodes them with a flow-matching (FM) decoder.
    \textbf{Stage 2:} an image encoder with learnable scene tokens integrates multi-view information into a unified scene latent space,
    supervised by the FM loss with the Stage-1 decoder frozen. 
    During \textbf{inference}, only the second stage pipeline is used to produce a complete, non–pixel-aligned point cloud.}
    \label{fig:pipeline}
    \vspace{-0.5cm}
\end{figure*}

\paragraph{Complete 3D Reconstruction.}
To achieve a complete 3D reconstruction,
existing approaches typically follow two main paradigms.
One line of work~\citep{vahdat2022lion,zhang20233dshape2vecset,zhao2023michelangelo,zhang2024clay,ren2024xcube,xiang2024structured,TripoSR2024,yang2024hunyuan3d,hunyuan3d22025tencent,li2025triposg} leverages compact latent spaces~\citep{rombach2022high} 
or large-scale networks~\citep{hong2023lrm,zhang2024gs,tang2025lgm}
for generating complete 3D assets.
While effective, these approaches primarily target individual \emph{object} reconstruction and fall short in modeling complex, cluttered scenes.
The other paradigm fine-tunes large-scale pre-trained diffusion models~\citep{rombach2022high,blattmann2023stable}.
For \emph{objects},
a notable example is Zero-1-to-3~\citep{liu2023zero},
which conditions on camera pose for high-quality 360-degree novel view rendering by training on a huge dataset, Objaverse~\citep{deitke2023objaverse}.
This is followed by a large group of successors~\citep{long2024wonder3d,shi2024mvdream,han2024vfusion3d,liu2023one,li2024instant3d,zheng2024free3d,ye2024consistent,voleti2025sv3d}.
For \emph{scenes},
several recent approaches aim to achieve complete 3D geometry by leveraging controlled camera trajectories~\citep{wang2024motionctrl,sargent2024zeronvs,wu2024reconfusion,gao2024cat3d,wallingford2024image,zhou2025stable} or introducing auxiliary conditioning signals~\citep{liu2024reconx,yu2024viewcrafter,chen2024mvsplat360,yu2025wonderworld}.
However, these methods do not explicitly reconstruct the complete underlying 3D geometry.
More recently,
WVD~\citep{zhang2024world} and Bolt3D~\citep{szymanowicz2025bolt3d} propose a hybrid RGB+point map representation to combine geometry and appearance for 3D reconstruction;
however, they still require known camera poses for novel RGB+point map rendering.
We address \emph{pose-free} 3D reconstruction from unconstrained images, and provide a complete 3D representation.
More closely related to our work,
Amodal3R~\citep{wu2025amodal3r} introduces amodal 3D reconstruction to reconstruct complete 3D assets from partially visible pixels,
but it still works only on objects.

\section{Method}
\label{sec:method}
Given a set of unposed images
$\mathcal{I}=\{{\bm I}^{i}\}_{i=1}^K$,
(${\bm I}^i \in \mathbb{R}^{H \times W \times 3}$) of a scene,
our goal is to learn a neural network
$\Phi$
that directly produces a complete 3D point cloud,
both in terms of 
\emph{visible} and \emph{occluded} regions.
We first discuss the problem
formulation in~\Cref{sec:method_formulation},
followed by our 3D latent autoencoder
in~\Cref{sec:method_encoder_decoder},
and we finally describe our global scene representation
in~\Cref{sec:method_scene_token}.

\subsection{Problem Formulation}
\label{sec:method_formulation}

\paragraph{Problem Definition.}

\begin{wrapfigure}{r}{0.5\textwidth} % {placement}{width}
  \vspace{-2\baselineskip}             % tweak vertical spacing (optional)
  \centering
% \begin{figure*}[tb!]
    % \centering
    \includegraphics[width=\linewidth]{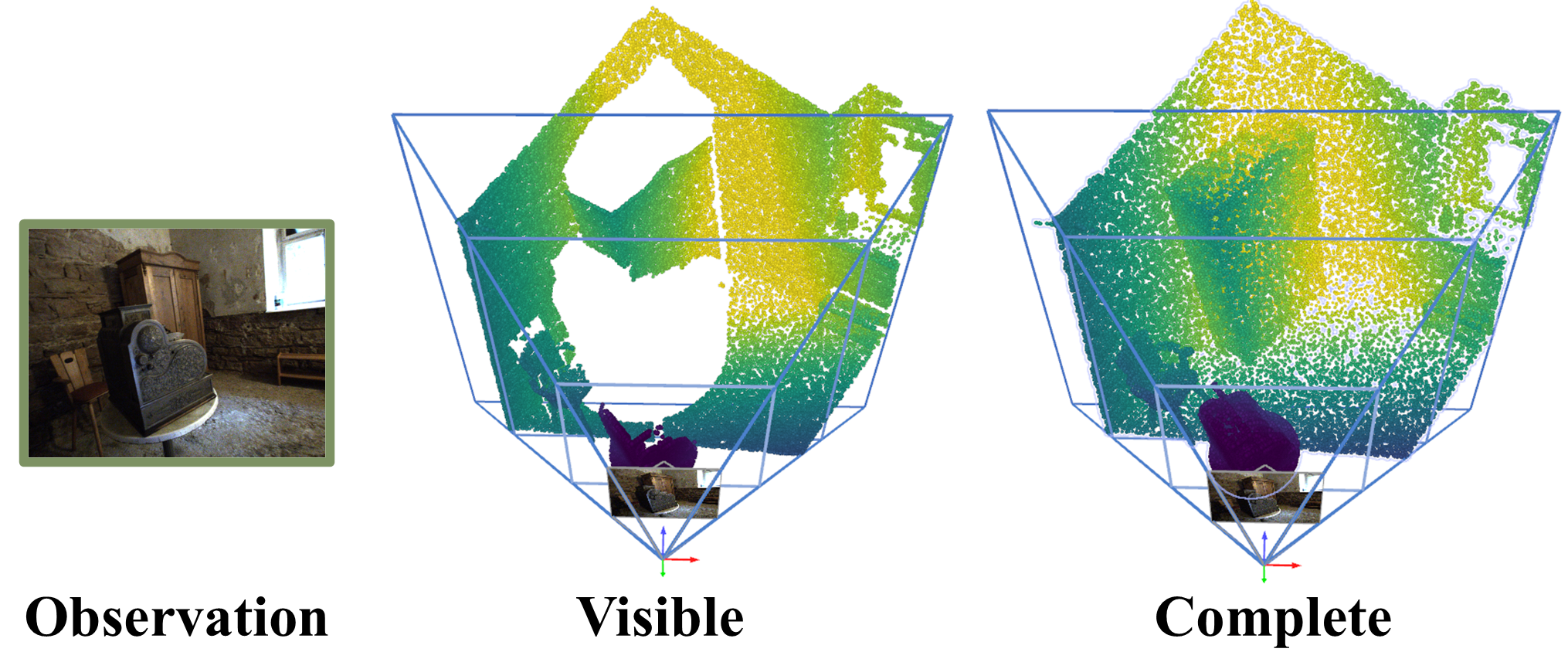}
    \caption{ \footnotesize \textbf{Visible point clouds \emph{vs.} complete point clouds.} Our \method~aims to recover the complete geometry within the input view's frustum.}%
    \label{fig:data_sampling}
% \end{figure*}
  \vspace{-1em}             % optional
\end{wrapfigure}

The input to our model is a set of $K$ \emph{unposed images}
$\mathcal{I}=\{{\bm I}^{i}\}_{i=1}^K$
of a scene,
and the output is a \emph{complete} 3D point cloud
$P\in\mathbb{R}^{N\times 3}$,
using a feed-forward neural network
$\Phi: \mathcal{I} \rightarrow P$.
This task is conceptually similar to the conventional feed-forward 3D reconstruction setting~\citep{wang2024dust3r,wang2025vggt,jiang2025rayzer},
except that here
$N$ represents the number of points in the \emph{complete} scene point cloud (as shown in~\Cref{fig:data_sampling}),
rather than
$K\times H\times W$
points back-projected from all pixels in each input image.

The \emph{key observation} is that a scene in the real world is composed of a fixed set of physical points,
regardless of how many images are captured from different viewpoints. If a physical 3D point is observed in multiple 2D images, the correct representation of the scene should contain a single point,
rather than duplicated points back-projected from each observation.
Conversely, even if a physical 3D point is never observed in any image, it still exists in the real world and should be inferred by the model.
Therefore, the model should be able to predict the occluded regions of the scene and avoid generating redundant points in the overlapping visible regions.

\paragraph{Data Preprocessing.}

The key to training such a model is the definition of the \emph{complete} 3D point clouds of a scene.
It must contain points in both \emph{visible} and \emph{occluded} regions,
and avoid duplicated points in the \emph{overlapping visible} regions.
The visibility of a 3D point is defined with respect to the input images $\mathcal{I}$.
However, the notion of invisible points is ambiguous:
there are infinitely many points that are not visible in input images,
or even outside the field of view of all input images.
To simplify the problem,
as shown in~\Cref{fig:data_sampling}, 
we define invisible points within the input-view frustum and discard points outside the frustum.

Creating such complete point clouds for supervision is non-trivial.
The ideal solution is to use the ground-truth 3D mesh of the scene,
which can be easily converted to a complete point cloud by uniformly sampling points on the mesh surface.
However, the ground-truth mesh is not always available in scene-level datasets.
When ground-truth 3D meshes are not available,
we instead approximate the complete point clouds using depth maps aggregated from dense views.
Specifically,
we first back-project the depth maps from all dense views into point clouds,
then apply voxel-grid filtering to remove duplicate points in overlapping visible regions.
Finally, we discard points outside the frustum of the selected input views (single, two, or a set of views).
During training, we apply the farthest point sampling method with random initialization to obtain a subset from the complete point cloud to train our point decoder.

Importantly,
as in DUSt3R~\citep{wang2024dust3r},
our complete point clouds are also \emph{view-agnostic}:
the 3D points are defined in the coordinate system of the first input view $\bm{I}_1$,
but are \emph{not} pixel-aligned to any input images.
This design allows the model to learn to reconstruct the complete 3D scene in the first view's coordinate system while ignoring the ambiguity of pose estimation.
Consequently,
our model can be trained on a wide range of datasets without requiring ground-truth meshes,
unlike existing object-level methods~\citep{zhang20233dshape2vecset,li2025triposg,yang2024hunyuan3d}.

\subsection{3D Latent Encoder-Decoder with Flow Matching}
\label{sec:method_encoder_decoder}
Following recent works in 3D latent vector representation~\citep{zhang20233dshape2vecset},
we design a 3D latent transformer~\citep{vaswani2017attention}.
However, ours does \emph{not} require a perfect mesh as input or supervision.
As shown in~\Cref{fig:pipeline} (Stage 1), we implement the model as a diffusion model.

\paragraph{Diffusion-based 3D AutoEncoder.}
The encoder $\Phi_{\text{enc}}$
takes the point cloud $P\in\mathbb{R}^{N\times 3}$ as input,
and outputs a set of $M$ latent tokens
$Z\in\mathbb{R}^{M\times C}$.
In practice, to reduce the computational cost, the initial query points
$q\in \mathbb{R}^{M\times 3}$
are sampled from the complete point cloud $P\in\mathcal{R}^{N\times 3}$ using farthest point sampling,
where $M \ll N$.
We further design a hybrid query representation by concatenating the point query with learnable tokens of the same length $M$ along the channel dimension,
followed by a linear projection layer that reduces the channel dimension from $2C$ to $C$.

Once the latent tokens $Z$ are obtained,
existing 3D VAE methods~\citep{zhang20233dshape2vecset,hunyuan3d22025tencent,li2025triposg} typically use a deterministic decoder to predict an occupancy field
$o=\Phi_{\text{dec}}(Z,x)$ or SDF values $s=\Phi_{\text{dec}}(Z,x)$ for each 3D grid query $x\in\mathbb{R}^{N\times3}$.
However, this is not suitable for our task,
since obtaining ground-truth occupancy or SDF values for real scene-level datasets is costly or even infeasible. 
Importantly,
unlike objects that can be enclosed within a canonical space,
scenes typically lack well-defined boundaries and expand as the number of observations increases,
making it difficult to predefine a canonical space.
Instead, we directly predict the 3D coordinates of each query point.
However, because point clouds are \emph{not} ordered or aligned,
we cannot directly map the 3D point query to the ground-truth point clouds $P$ using an $L_2$ loss.
We then adopt a diffusion-based decoder $\Phi_{\text{dec}}(x_t,Z,t)$ to decode the scene tokens $Z$ back to the original point cloud space.
% producing $\hat{P}\in\mathbb{R}^{N\times 3}$.
The transformer-based decoder takes as input a set of $N$ noised query point clouds
$x_t\in\mathbb{R}^{N\times 3}$,
at the flow matching time $t$,
and the latent tokens $Z$ as conditioning. 
The whole architecture is trained end-to-end with a flow matching loss~\citep{lipman2023flow}:
\begin{equation}
    \mathcal{L}^{\text{AE}}_{\text{flow}} = 
    \mathbb{E}_{t,x_0\sim P,\epsilon\sim\mathcal{U}(-1,1)} 
    \left[ \left\| \Phi_{\text{dec}}(x_t,Z,t) - (\epsilon - x_0) \right\|_2^2 \right],
\end{equation}
where
$x_t = (1-t) x_0 + t\epsilon$.
Note that we do \emph{not} use KL loss or any other regularization on the latent tokens as in existing 3D latent VAE methods~\citep{hunyuan3d22025tencent,li2025triposg}.

\begin{figure*}[tb!]
    \centering
    \includegraphics[width=0.95\linewidth]{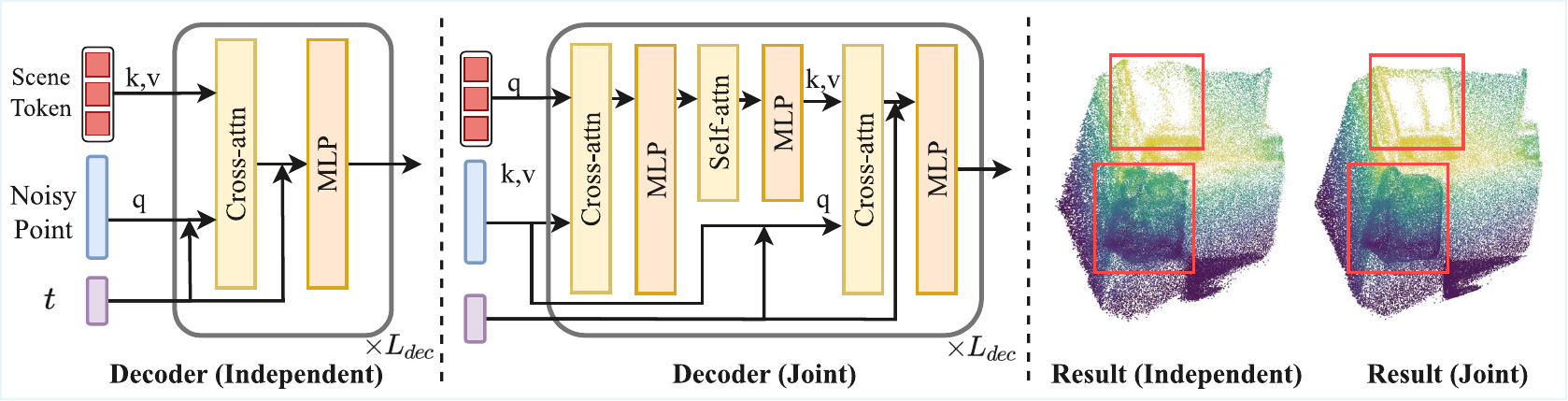}
    \vspace{-10pt}
    \caption{\textbf{Different Decoder Architectures.}
    The independent decoder uses cross-attention only, while the joint decoder implements an efficient self-attention, which yields more precise structures. }%
    \label{fig:comparison_decoder}
    \vspace{-0.5cm}
\end{figure*}

\paragraph{Architecture.}
As noted above,
our 3D autoencoder is implemented with a transformer architecture.
Specifically,
the encoder is built upon TripoSG~\citep{li2025triposg},
which consists of one cross-attention layer and eight self-attention layers.
The decoder has three transformer blocks (details are shown in~\Cref{fig:comparison_decoder}). 
% two cross-attention layers and one self-attention layer in each block.
Notably, the query will be switched between the 3D latent tokens
$Z$ and the noisy point clouds $x_t$ in each cross-attention layer.
This design reduces the size of the self-attention maps while preserving information flow between latent tokens and query points.
Concurrent work~\citep{chang20243d} also proposes a diffusion-based 3D latent autoencoder, but they consider a 3D shape as a probability density function,
and process each point independently.

\subsection{Scene Representation with Learnable Tokens}
\label{sec:method_scene_token}
We now describe how to learn a global scene representation
from a set of unposed images.
As shown in~\Cref{fig:pipeline} (Stage 2), we implement it using a large transformer that takes the input images
$\mathcal{I}$ and a set of $M$ learnable tokens
$t_{S}\in\mathbb{R}^{M\times C}$ as input,
and outputs the scene representation
$\hat{Z}\in\mathbb{R}^{M\times C}$.

\paragraph{Learnable Scene Tokens.}
As mentioned in~\Cref{sec:method_formulation},
our model aims to predict a \emph{fixed} number of \emph{non-pixel-aligned} points under the first view's coordinate system.
Accordingly,
in addition to $L$ patchified image tokens
$t_I\in\mathbb{R}^{L\times C}$, we introduce a set of $M$ learnable global scene tokens $t_{S}\in\mathbb{R}^{M\times C}$,
which are randomly initialized and optimized during training.
The combined token set $t_I \cup t_S$ from all input images,
\emph{i.e.}, $t_I=\cup_{i=1}^K\{t_I^i\}$,
and the learnable scene tokens $t_{S}$, is fed into a large transformer,
with multiple frame- and global-level self-attention layers.
To simplify the architecture, the learnable scene tokens
$t_{S}$ are treated as a global frame underlying the first view's coordinate system.
This means that the scene tokens undergo the same operations as the image tokens in each Transformer block, except that they use the first view’s camera token.

\paragraph{Architecture.}
Our image encoder is built upon VGGT~\citep{wang2025vggt}, a representative feed-forward 3D reconstruction model.
However, we do not use its dense prediction heads
% of VGGT
to predict the \emph{pixel-aligned} depth and point maps.
Instead, we use the output scene tokens
$\hat{Z}\in\mathbb{R}^{M\times C}$
as the conditioning of our point decoder
$\Phi_{\text{dec}}$,
to predict the \emph{non-pixel-aligned} complete 3D point clouds
$\hat{P}\in\mathbb{R}^{N\times 3}$.
The entire architecture is trained end-to-end with the flow matching loss: 
\begin{equation}
    \mathcal{L}^{\text{Tran}}_{\text{flow}} = 
    \mathbb{E}_{t,x_0\sim P,\epsilon\sim\mathcal{U}(-1,1)} 
    \left[ \left\| \Phi_{\text{dec}}(x_t,\hat{Z},t) - (\epsilon - x_0) \right\|_2^2 \right],
\end{equation}
where $\Phi_\text{dec}$ is frozen in Stage 2,
and only the transformer $\Phi_\text{tran}: t_I\cup t_S\to \hat{Z}$
and the learnable scene tokens $t_{S}$ are optimized.
\section{Experiments}
\label{sec:experiments}

\subsection{Experimental Settings}
\label{sec:experimental_setting}

\paragraph{Metrics.}
Following~\citet{li2025lari},
we report Chamfer Distance (CD) and F-score (FS) at different thresholds (\emph{e.g.,} 0.1, 0.05) for completion tasks.
For multi-view reconstruction tasks,
we report accuracy (Acc), completion (Comp), and normal consistency (NC) following~\citet{wang2025continuous}. Best results are highlighted as \colorbox{colorFst}{\bf first}, \colorbox{colorSnd}{second}, and \colorbox{colorTrd}{third}.

\begin{table}[tb!]
\centering
\footnotesize
% \scriptsize
\caption{
\textbf{Quantitative results for scene completion on SCRREAM}~\citep{jung2024scrream}.
The \emph{one-side} Chamfer Distance (GT $\rightarrow$ Prediction) results are shown in (~).
$K$ is the number of input views.
$*$ denotes methods that are not trained on scene-level data.
Our method shows better completion results compared to other competitive baselines.
Note that,
since \method~is a \emph{non-pixel-aligned} 3D reconstruction model,
it does not explicitly distinguish the visible and occluded points. 
}
\label{tab:scrream}
\setlength\tabcolsep{2pt}
\resizebox{0.9\linewidth}{!}{
\begin{tabular}{@{}llccccccccc@{}}
\toprule
\multirow{2}{*}{\textbf{Type}} &
\multirow{2}{*}{\textbf{Method}} &
\multicolumn{3}{c}{\lo\textbf{Visible ($K$=1)}} &
\multicolumn{3}{c}{\textbf{Complete ($K$=1)}} &
\multicolumn{3}{c}{\textbf{Complete ($K$=2)}} \\
\cmidrule(l){3-5} \cmidrule(l){6-8} \cmidrule(l){9-11}
& & \lo CD$\downarrow$ &\lo FS@0.1$\uparrow$ &\lo FS@0.05$\uparrow$
  & CD$\downarrow$ & FS@0.1$\uparrow$ & FS@0.05$\uparrow$
  & CD$\downarrow$ & FS@0.1$\uparrow$ & FS@0.05$\uparrow$ \\
\midrule
\multirow{2}{*}{Object}
& TripoSG*     & \lo(0.268) & \lo(0.418) & \lo(0.301) & 0.242 & 0.467 & 0.333 & -    & -     & -     \\
& TRELLIS*     & \lo(0.301) & \lo(0.420) & \lo(0.313) & 0.256 & 0.429 & 0.312 & 0.286 & 0.402 & 0.288 \\
\midrule
\multirow{5}{*}{\shortstack[l]{Single-\\view}}
& Metric3D-v2  & \lo0.063 & \lo0.803 & \lo0.534 & 0.086 & 0.725 & 0.473 & - & - & - \\
& DepthPro     & \lo0.055 & \lo0.852 & \lo0.603 & 0.079 & 0.764 & 0.535 & - & - & - \\
& MoGe         & \lo\textbf{0.035} & \lo\textbf{0.945} & \lo\textbf{0.786} & \trd 0.063 & \nd0.836 & \nd0.668 & - & - & - \\
& LaRI    & \lo0.057 & \lo0.847 & \lo0.589 & \nd0.059 & \trd0.825 & 0.590 & - & - & - \\
\midrule
\multirow{4}{*}{\shortstack[l]{Multi-\\view}}
& DUST3R       & \lo0.059 & \lo0.851 & \lo0.653 & 0.086 & 0.757 & 0.565 & \nd0.061 & \nd0.833 & \nd0.641 \\
& CUT3R        &\lo 0.069 &\lo 0.835 &\lo 0.679 & 0.091 & 0.753 & 0.543 & 0.092 & 0.739 & 0.532 \\
& VGGT         &\lo 0.041 &\lo 0.923 &\lo 0.754 & 0.070 & 0.810 & \trd0.657 & \trd0.065 & \trd0.821 & \trd0.606 \\
& Ours         &\lo (0.043) &\lo (0.904) &\lo (0.730) & \st 0.048 &\st 0.882 & \st0.687 & \st0.053 & \st0.862 &\st0.657 \\
\bottomrule
\end{tabular}
}
\vspace{-0.5em}
\end{table}

\paragraph{Implementation Details.}
By default, we set the number of scene tokens as $M=768$ and the number of points as
$N=10,000$
for training.
The image encoder architecture is exactly the same as VGGT~\citep{wang2025vggt},
while the 3D latent autoencoder contains 8 layers in the encoder and 3 layers in the decoder.
The training contains two stages.
In Stage 1, we train the autoencoder for 50 epochs.
In Stage 2, we initialize the image encoder with VGGT pretrained weights and the flow-matching decoder with Stage-1 weights, then train for another 50 epochs.
Note that,
we only fine-tune the image encoder and the scene-token transformer in Stage 2.
We train both stages by optimizing the flow-matching loss with the AdamW optimizer and a learning rate of 3e-4.
The training runs on 4 NVIDIA A40 GPUs with a total batch size of 32. \rebuttal{We use standard flow-matching with cosine noise scheduling, timestep sampling in [0,1], a fixed 0.04 step size at inference, and identical loss settings for both object-level and scene-level datasets.}

\begin{figure*}[tb!]
    \centering
    \vspace{-2em}
    \includegraphics[width=0.85\linewidth]{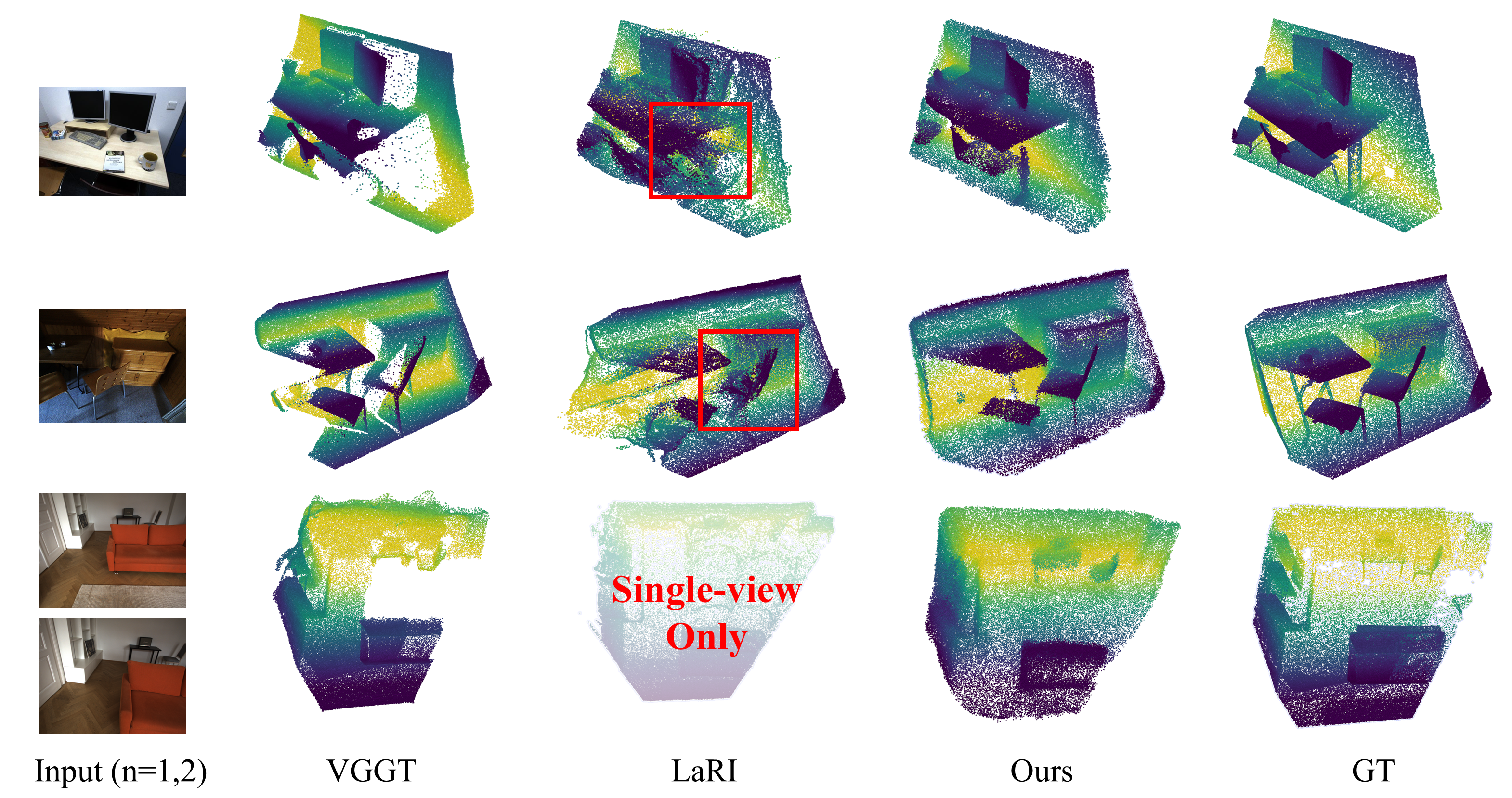}
    \caption{\textbf{Qualitative results for scene completion on SCRREAM}~\citep{jung2024scrream}. Our method produces more complete point clouds with clearer and less distorted geometry than other baselines. }%
    \vspace{-1em}
    \label{fig:scrream}
\end{figure*}

\begin{table}[t]
\footnotesize
\centering\
\caption{\rebuttal{\textbf{Quantitative results for hole area ratio and point cloud density variance on SCRREAM~\citep{jung2024scrream}.} Our method significantly outperforms pixel-aligned baselines, achieving both lower hole ratios and lower density variance.} }
\label{tab:scene_hole_density}
\setlength\tabcolsep{6pt}
\resizebox{0.9\linewidth}{!}{
\begin{tabular}{@{}lcccccc@{}}
\toprule
\multirow{2}{*}{Method} & \multicolumn{2}{c}{\textbf{Complete} (K=1)} & \multicolumn{2}{c}{\textbf{Complete} (K=2)} & \multicolumn{2}{c}{\textbf{Complete} (K=4)}\\ \cmidrule(l){2-7} 
                        & Hole Ratio$\downarrow$    & Density Var. $\downarrow$  & Hole Ratio $\downarrow$    & Density Var. $\downarrow$  & Hole Ratio $\downarrow$    & Density Var. $\downarrow$   \\ \midrule
DUST3R  & \trd{0.317} & \trd{7.758} & \trd{0.237} & \trd6.553 & \nd{0.257} & \trd 4.801 \\
CUT3R   & 0.363 & 8.402 & \nd{0.237} & 6.554 & 0.326 & \nd{4.658} \\
VGGT    & \nd{0.307} & \nd{7.105} & 0.238 & \nd{6.546} & \trd 0.261 & 5.217 \\
Ours&  \st 0.088 & \st 5.127  &  \st 0.121  & \st 2.188  & \st 0.134 & \st 1.881 \\ \bottomrule
\end{tabular}}
\vspace{-1em}
\end{table}

\subsection{Scene-level Reconstruction}

\paragraph{Datasets.}
The scene-level model was trained on 3D-FRONT~\citep{fu20213d} and ScanNet++V2~\citep{yeshwanth2023scannet++}, using the training splits from LaRI~\citep{li2025lari} and DUSt3R~\citep{wang2024dust3r}, which contain 100k and 230k unique images, respectively. For visible part training, we further incorporate ARKitScenes~\citep{dehghan2021arkitscenes}. 
Ideally,
our model is able to handle an arbitrary number of input views, similar to VGGT~\citep{wang2025vggt}.
However,
limited by the available computational resources,
we mainly verify our contributions on two-view pairs and train with 1–2 input views.

To evaluate the cross-domain generalization ability of our model,
we directly evaluate performance on the unseen SCRREAM dataset~\citep{jung2024scrream},
which provides complete ground-truth scans.
We follow LaRI's setting for single-view evaluation, with 460 images for testing.
\rebuttal{For the two-view setting, we sample 329 pairs from the same scene with a frame-ID distance of 40–80, where the maximum pose gap is 30\% (measured by point cloud co-visibility) and the hole area ratios (measured by completeness with threshold 0.1) range from 5.3\% to 48.6\%.}
We additionally evaluate visible-surface multi-view reconstruction on the 7-Scenes~\citep{shotton2013scene} and NRGBD datasets~\citep{azinovic2022neural},
sampling input images at intervals of 100 frames.

\paragraph{Baselines.}
We compare \method~with several representative scene-level 3D reconstruction methods,
including \textbf{i)} single-view Metric3D-v2~\citep{hu2024metric3d}, DepthPro~\citep{bochkovskii2024depth}, and MoGe~\citep{wang2025moge};
\textbf{ii)} multi-view DUSt3R~\citep{wang2024dust3r}, CUT3R~\citep{wang2025continuous}, and VGGT~\citep{wang2025vggt}.
However, these methods only focus on \emph{pixel-aligned visible} 3D reconstruction.
Hence, we further compare with the concurrent complete 3D reconstruction work LaRI~\citep{li2025lari}.
Since it does not support multi-view inputs,
for completeness,
we also report the results from object-level methods TripoSG~\citep{li2025triposg} and TRELLIS~\citep{xiang2025structured} by disabling the input mask,
while they are not trained on scene-level data.

\begin{figure*}[tb!]
    \centering
    \includegraphics[width=\linewidth]{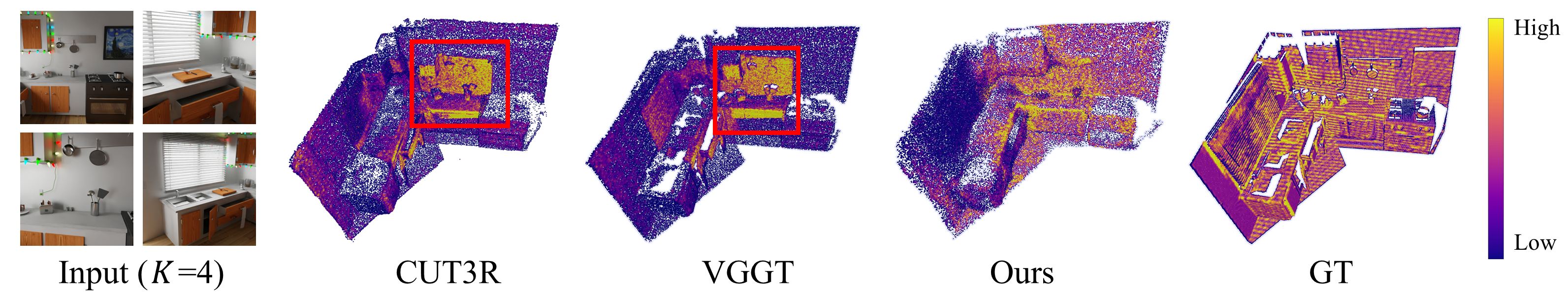}
    \vspace{-20pt}
    \caption{\textbf{Qualitative results for density evaluation on NRGBD ($K=4$)~\citep{azinovic2022neural}.}
    Yellow regions denote higher density, and purple regions denote lower density. Despite being trained with only two views, \method~generalizes well to multiple views ($K=4$).
}%
    \label{fig:nrgbd_vis}
\end{figure*}

\begin{table}[tb!]
\vspace{-10pt}
\centering
\footnotesize
\caption{\textbf{Quantitative results on visible reconstruction on 7-Scenes ($K$=2)~\citep{shotton2013scene}.}
 Our \method~model can be trained on RGB-D data and achieves competitive results compared to multi-view reconstruction methods.
 Note that, we use fewer tokens to represent a 3D scene.
 }
\label{tab:7scenes_nrgbd}
\setlength\tabcolsep{12pt}
\resizebox{0.85\linewidth}{!}{
\begin{tabular}{@{}lcccccccc@{}}
\toprule
\multirow{2}{*}{\textbf{Method}} &
\multirow{2}{*}{\textbf{\# Tokens}} &
 \multicolumn{2}{c}{Acc $\downarrow$} &
      \multicolumn{2}{c}{Comp $\downarrow$} &
      \multicolumn{2}{c}{NC $\uparrow$} \\
\cmidrule(lr){3-4}\cmidrule(lr){5-6}\cmidrule(lr){7-8}
& & Mean & Med. & Mean & Med. & Mean & Med. \\
\midrule
DUSt3R      & 2048 & 0.054 & 0.023 & 0.075 & 0.034 & 0.772 & 0.901 \\
Spann3R        & 784  & 0.044 & \trd0.022 & \trd0.046 & \nd0.025 & \trd0.792 & \nd0.922 \\
CUT3R          & 768  & \trd0.043 & 0.023 & 0.054 & 0.028 & 0.760 & 0.884 \\
VGGT           & 2738 & \nd0.042 & \st{0.020} & \nd0.045 & \nd0.025 & \st{ 0.813} & \st{0.923} \\
Ours           & 768  & \st{0.041} & \nd0.021 & \st{0.033} & \st{0.019} & \nd0.794 & \trd0.917 \\
\bottomrule
\end{tabular}
}
\vspace{-1em}
\end{table}

\paragraph{Scene Completion.}
Following LaRI,
we evaluate our amodal 3D reconstruction results on both \emph{visible} and \emph{complete (visible + occluded)} regions. 
\rebuttal{For visible setting, we follow the same evaluation protocol as DUST3R~\citep{wang2024dust3r} and VGGT~\citep{wang2025vggt}, where the ground truth contains only the visible points from the input views. For the complete setting, we use the full point cloud as ground truth, including occluded and unseen regions.}
However, unlike pixel ray-conditional LaRI,
\method~does not explicitly identify the visible region.
We therefore adopt \emph{one-sided} Chamfer Distance (GT $\rightarrow$ Prediction) for the visible part:
each GT-visible point must be explained by a nearby prediction.
This measures coverage of the visible ground truth,
yet without penalizing missing, occluded regions.
\Cref{tab:scrream} shows three settings:
1-view \emph{visible},
1-view \emph{complete},
and 2-view \emph{complete}. 
Despite using only two datasets to train,
% \cxz{why do we use less training data, lari also uses just two datasets?},
our method consistently outperforms multi-view baselines on complete reconstruction in both $K=1$ and $K=2$ settings,
demonstrating the effectiveness of our non–pixel-aligned approach.
Our method also achieves competitive results on visible-surface reconstruction.
Qualitative results in \Cref{fig:scrream} show that our method produces surfaces without holes (unlike pixel-aligned methods such as VGGT) and yields clearer,
less distorted geometry than LaRI.
This benefit is attributed to our non–pixel-aligned design,
which prevents ray-direction bias in reconstruction.
\rebuttal{We further quantify the completion capability using the hole area ratio, which is computed by checking whether each ground-truth point has a predicted point within a distance threshold of 0.1. As shown in~\Cref{tab:scene_hole_density}, our method consistently achieves significantly lower hole ratios, demonstrating its strong capability for complete reconstruction. In terms of density variance, our approach outperforms all pixel-aligned baselines, even in unseen four-view settings, indicating better physical plausibility with more evenly distributed point clouds. 
Moreover, when comparing across different $K$, the density variance consistently decreases from one to four input views, further confirming that incorporating more views leads to improved spatial uniformity.
}

\paragraph{Physically-plausible Reconstruction.}
Beyond 3D completion,
our \emph{non–pixel-aligned} formulation also features physically plausible reconstruction by fusing evidence in 3D rather than along camera-pixel rays,
reducing duplicated points in overlapping regions and improving cross-view consistency.
To illustrate this,
we evaluate visible reconstruction with $K=4$ views on NRGBD~\citep{azinovic2022neural}.
% \cxz{I still think it is not so necessary to test 4 view.}
As shown in \Cref{fig:nrgbd_vis},
pixel-aligned methods like CUT3R~\citep{wang2025continuous} and VGGT~\citep{wang2025vggt} accumulate 3D points in co-visible regions,
producing uneven densities and multi-layer artifacts.
This is physically incorrect,
as each point corresponds to a single location in the real world,
regardless of the number of views.
In contrast,
our \method~generates cleaner,
single-surface geometry with uniform point distribution,
achieving competitive results despite using fewer datasets and views (see \Cref{tab:7scenes_nrgbd}). 
We further quantify physical plausibility by computing the density variance in~\Cref{tab:scene_hole_density}, which indicates that our method achieves a more uniformly distributed reconstruction compared to pixel-aligned baselines.

\begin{table}[tb!]
\centering
\small
\caption{\textbf{Quantitative results for object completion on GSO~\citep{downs2022google}.}
\method~provides a unified solution for both scene and object completion from unposed images.
}
\label{tab:gso}
\setlength\tabcolsep{7pt}
\resizebox{0.9\linewidth}{!}{
\begin{tabular}{llcccccc}
\toprule
\multirow{2}{*}{\textbf{Type}} &
\multirow{2}{*}{\textbf{Method}} &
\multicolumn{3}{c}{\textbf{View-aligned ($K$=1)}} &
\multicolumn{3}{c}{\textbf{View-aligned ($K$=2)}} \\
\cmidrule(l){3-5}\cmidrule(l){6-8}
& & CD $\downarrow$ & FS@0.1 $\uparrow$ & FS@0.05 $\uparrow$
  & CD $\downarrow$ & FS@0.1 $\uparrow$ & FS@0.05 $\uparrow$ \\
\midrule
\multirow{5}{*}{\shortstack[l]{Single-\\view}}
& SF3D           & 0.037 & 0.913 & 0.738 & -     & -     & -     \\
& SPAR3D         & 0.038 & 0.912 & 0.745 & -     & -     & -     \\
& LaRI     & \nd0.025 & \nd0.966 & 0.894 & -     & -     & -     \\
& TripoSG        & \nd 0.025 & 0.961 & \nd0.899 & -     & -     & -     \\
\midrule
\multirow{2}{*}{\shortstack[l]{Multi-\\view}}
& TRELLIS        & \nd 0.025 & \trd0.962 & \trd0.896 & \nd0.028 &\nd 0.946 &\nd 0.874 \\
& Ours     & \st{0.020} & \st{0.985} & \st0.925 & \st{0.023} & \st{0.978} & \st{0.903} \\
\bottomrule
\end{tabular}
}
\vspace{-0.5em}
\end{table}

\begin{figure*}[tb!]
    \centering
    \includegraphics[width=0.9\linewidth]{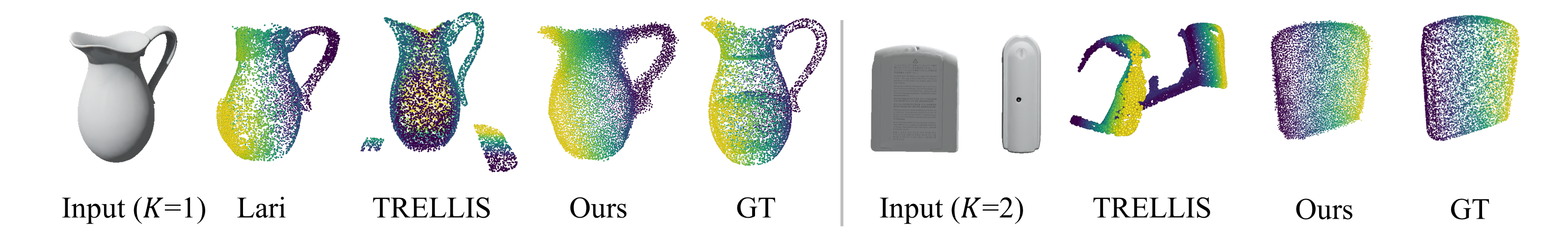}
    \caption{\textbf{Qualitative results for object completion on GSO~\citep{downs2022google}.}
    Our method provides more precise geometry and better 3D consistency with multi-view inputs.}%
    \label{fig:gso}
    \vspace{-15pt}
\end{figure*}

\subsection{Object-Level Reconstruction} 

\paragraph{Datasets.}
We demonstrate the versatility of our method as a unified non–pixel-aligned approach for both scenes and objects.
Following \citet{li2025lari},
we train an object-completion model on Objaverse~\citep{deitke2023objaverse} with 190k annotated images.
For evaluation,
we report results on unseen Google Scanned Objects~\citep{downs2022google}.
For single-view reconstruction,
we use the same 1030-object split as LaRI~\citep{li2025lari}.
For two-view reconstruction,
we fix the 0th image and uniformly sample three additional views,
yielding three pairs per object (3090 pairs in total).

\paragraph{Baselines.}
We compare with several representative object-level 3D reconstruction methods,
including SF3D~\citep{boss2025sf3d}, SPAR3D~\citep{huang2025spar3d}, TripoSG~\citep{li2025triposg}, and TRELLIS~\citep{xiang2025structured}.
We also include LaRI~\citep{li2025lari} as a strong baseline,
which is trained on the same dataset and supports amodal 3D reconstruction.

\paragraph{Object Completion.}
Table~\ref{tab:gso} reports results for single view ($K=1$) and two views ($K=2$).
Our \method~outperforms LaRI on all three metrics.
Importantly,
our pipeline supports multi-view completion that maps different unposed images into the same view-aligned space.
On the multi-view benchmark,
our method also outperforms TRELLIS,
highlighting the benefits of non–pixel-aligned reconstruction for consistent global geometry.
Qualitative comparisons in Figure~\ref{fig:gso} show that our completions preserve fine structures,
and achieve better 3D consistency in the multi-view setting.

\begin{table*}[t]
\caption{\textbf{Ablations.}
All models are evaluated on the SCRREAM complete ($K=1$) setting. 
We report CD$\downarrow$, FS@0.1$\uparrow$, FS@0.05$\uparrow$ and FS@0.02$\uparrow$ across different ablation settings.}
\label{tab:ablations}
\centering
\footnotesize
\setlength{\tabcolsep}{4pt}
\resizebox{\linewidth}{!}{
\begin{tabular}{c|ccc|ccc|cc|cc}
\toprule
  & \multicolumn{3}{c|}{\textbf{Initial tokens (Stage 1)}} 
  & \multicolumn{3}{c|}{\textbf{\# Scene tokens (Stage 1)}} 
  & \multicolumn{2}{c|}{\textbf{FM Decoder (Stage 1)}} 
  & \multicolumn{2}{c}{\textbf{Img Resolution (Stage 2)}} \\
\midrule
  \textbf{Settings}& Point & Learnable & \bf Hybrid & 256 & 512 & \bf768 & Indep. & \bf Joint & 224 & \bf 518 \\
\midrule
CD$\downarrow$      & 0.011 & 0.013 & \bf 0.011 & 0.014 & 0.013 & \bf0.011 & 0.012 &\bf 0.011 & 0.054 & \bf0.048 \\
FS@0.1$\uparrow$    &  0.999 & 0.998 & \bf0.999 & 0.996 & 0.998 & \bf0.999 & 0.998 & \bf0.999 & 0.861 & \bf0.882 \\
FS@0.05$\uparrow$   & 0.991 & 0.981 & \bf 0.993 & 0.975 & 0.986 & \bf0.993 & 0.990 & \bf 0.993 & 0.648 & \bf0.687 \\
FS@0.02$\uparrow$   & 0.894 & 0.841 & \bf 0.904 & 0.811 & 0.839 & \bf0.904 & 0.889 & \bf 0.904 & 0.327 &\bf 0.350 \\
\bottomrule
\end{tabular}
}
\vspace{-10pt}
\end{table*}

\vspace{-1.0em}
\subsection{Ablation Studies}

\rebuttal{We perform comprehensive ablation studies on the SCRREAM complete ($K=1$) setting to validate the key design choices of our method, with particular emphasis on assessing the contribution of Scene Tokens to global structure modeling.} The results are summarized in \Cref{tab:ablations}, and we discuss each component in detail below.

\paragraph{Initial Query (Stage 1).}
Prior work~\citep{zhang20233dshape2vecset} shows that the initialization of point queries affects autoencoder performance. 
We compare three options:
(i) \emph{downsampled input points},
(ii) \emph{learnable query tokens},
and (iii) a \emph{hybrid} that concatenates (i) and (ii). 
Downsampled points preserve the input geometry distribution,
whereas learnable tokens add flexibility under source–target shifts. 
As shown in \Cref{tab:ablations},
the hybrid combines these benefits and yields the best results.

\paragraph{Number of latent scene tokens (Stage 1).}
As described in \Cref{sec:method},
we represent each scene with a fixed-length set of latent tokens.
The number of tokens $M$ directly affects the representation capacity and ability to capture fine details,
especially in large scenes.
We evaluate different numbers of scene tokens from \{256, 512, 768\}
and observe consistent improvements as the count increases (see \Cref{tab:ablations}).
To balance accuracy and efficiency, we use $M=768$ tokens by default.
Ideally,
$M$ could be further increased for better performance.
We leave this for other works to explore.

\paragraph{Different architecture of flow-matching decoder (Stage 1).}
The latest work \citep{chang20243d} also presents a flow-matching decoder for a point cloud encoder,
but it assumes that all points are independent (shown in~\Cref{fig:comparison_decoder}).
This design is efficient, but ignores spatial correlations between points.
In our work,
we instead adopt a lightweight \emph{self-attention + cross-attention} decoder that jointly reasons over points and scene tokens,
allowing information exchange across the point set. 
To investigate the effect of this design, we compare it with an independent variant without self-attention. Empirically, the joint decoder yields lower CD, higher F-scores, and sharper fine details (\Cref{tab:ablations}), with small quantitative but significant qualitative gains (\Cref{fig:comparison_decoder}).

\paragraph{Input image resolution (Stage 2).}
In Stage 2 (image-to-point),
we adopt a transformer to integrate information between image and scene tokens. 
The input resolution determines the number of image tokens used in the aggregation process.
With patch size 14,
a resolution of $224\times 224$ yields $16\times16=256$ tokens,
while a resolution of $518\times 518$ yields $37\times37=1369$ tokens. 
As shown in \Cref{tab:ablations},
training with $518\times 518$ resolution consistently improves CD and F-scores.

\begin{table}[]
\caption{\rebuttal{\textbf{Ablations on different training loss functions.}
All models are evaluated on the SCRREAM complete ($K=1$) setting. 
We report CD$\downarrow$, FS@0.1$\uparrow$, FS@0.05$\uparrow$ and FS@0.02$\uparrow$ and inference time$\downarrow$ for the decoder.}}
\label{tab:ablations_loss}
\centering
\footnotesize
\setlength{\tabcolsep}{4pt}
\begin{tabular}{@{}cccccc@{}}
\toprule
\multirow{2}{*}{\textbf{Training Loss}} & \multicolumn{5}{c}{\textbf{SCRREAM (Stage 1)}}               \\ \cmidrule(l){2-6} 
                      & CD $\downarrow$   & FS@0.1 $\uparrow$ & FS@0.05 $\uparrow$  & FS@0.02 $\uparrow$ & Inference Time (s) $\downarrow$ \\ \midrule
Chamfer distance      & 0.024 & 0.981  & 0.907   & 0.575   & \textbf{0.557 }         \\
Flow-matching         & \textbf{0.011} & \textbf{0.999}  & \textbf{0.993}   & \textbf{0.904}   & 2.985          \\ \bottomrule
\end{tabular}
\vspace{-1em}
\end{table}

\paragraph{Flow-matching loss vs. Chamfer distance loss.}
\rebuttal{To verify the necessity of flow-matching for unordered point cloud encoding, we conduct an ablation using the same architecture but replacing the flow-matching loss with Chamfer Distance. Both models were trained on SCRREAM (Stage 1) under the same protocol. As shown in~\Cref{tab:ablations_loss}, flow-matching achieves significantly better reconstruction quality and generalization. Chamfer Distance struggles in scene-level settings because its nearest-neighbor formulation is computationally expensive, sensitive to density imbalance, and unable to capture global structure across varying scales and input views, while flow-matching produces stable, complete, and globally consistent reconstructions.} 

\vspace{-0.5em}
\section{Conclusion}
\label{sec:conclusion}

We present \method,
a non-pixel-aligned framework for amodal 3D scene reconstruction from unposed images.
Unlike prior pixel-aligned methods,
our \method~achieves state-of-the-art results in amodal 3D reconstruction, including both visible and invisible points, on both scene and object levels.
Notably,
it pioneers a new paradigm for physically plausible scene reconstruction that reconstructs a uniform point cloud for the entire scene,
without holes or duplicated points.
This simple yet effective design makes it a promising solution for real-world applications.

\paragraph{Limitations and Discussion.}
Due to computational constraints,
we train our model with a relatively small number of scene tokens and point clouds and a moderate number of input views (up to 2).
Hence, the reconstruction quality may degrade for large-scale scenes with complex structures.
Future work can explore scaling up the model and training data to enhance performance and generalization.
In addition,
our model currently focuses on reconstructing static scenes and does not handle dynamic objects or temporal consistency across frames.

\subsubsection*{Acknowledgments}
We would like to thank Ruining Li, Zeren Jiang, and Brandon Smart for their insightful feedback on the draft. This work was supported by the ERC Advanced Grant “SIMULACRON” (agreement \#884679), the GNI Project “AI4Twinning”, and the DFG project CR 250/26-1 “4DYoutube". Chuanxia Zheng is supported by NTU SUG-NAP and National Research Foundation, Singapore, under its NRF Fellowship Award NRF-NRFF172025-0009. 

% \newpage
\bibliographystyle{iclr2026_conference}
\bibliography{iclr2026_conference}

@article{li2025lari,
  title={Lari: Layered ray intersections for single-view 3d geometric reasoning},
  author={Li, Rui and Zhang, Biao and Li, Zhenyu and Tombari, Federico and Wonka, Peter},
  journal={arXiv preprint arXiv:2504.18424},
  year={2025}
}

@article{chang20243d,
  title={3D Shape Tokenization via Latent Flow Matching},
  author={Chang, Jen-Hao Rick and Wang, Yuyang and Martin, Miguel Angel Bautista and Gu, Jiatao and Zhao, Xiaoming and Susskind, Josh and Tuzel, Oncel},
  journal={arXiv preprint arXiv:2412.15618},
  year={2024}
}

@article{jung2024scrream,
  title={Scrream: Scan, register, render and map: A framework for annotating accurate and dense 3d indoor scenes with a benchmark},
  author={Jung, HyunJun and Li, Weihang and Wu, Shun-Cheng and Bittner, William and Brasch, Nikolas and Song, Jifei and P{\'e}rez-Pellitero, Eduardo and Zhang, Zhensong and Moreau, Arthur and Navab, Nassir and others},
  journal={Advances in Neural Information Processing Systems (NeurIPS)},
  volume={37},
  pages={44164--44176},
  year={2024}
}

@inproceedings{downs2022google,
  title={Google scanned objects: A high-quality dataset of 3d scanned household items},
  author={Downs, Laura and Francis, Anthony and Koenig, Nate and Kinman, Brandon and Hickman, Ryan and Reymann, Krista and McHugh, Thomas B and Vanhoucke, Vincent},
  booktitle={International Conference on Robotics and Automation (ICRA)},
  pages={2553--2560},
  year={2022},
  organization={IEEE}
}

@inproceedings{fu20213d,
  title={3d-front: 3d furnished rooms with layouts and semantics},
  author={Fu, Huan and Cai, Bowen and Gao, Lin and Zhang, Ling-Xiao and Wang, Jiaming and Li, Cao and Zeng, Qixun and Sun, Chengyue and Jia, Rongfei and Zhao, Binqiang and others},
  booktitle={International Conference on Computer Vision (ICCV)},
  pages={10933--10942},
  year={2021}
}

@inproceedings{yeshwanth2023scannet++,
  title={Scannet++: A high-fidelity dataset of 3d indoor scenes},
  author={Yeshwanth, Chandan and Liu, Yueh-Cheng and Nie{\ss}ner, Matthias and Dai, Angela},
  booktitle={International Conference on Computer Vision (ICCV)},
  pages={12--22},
  year={2023}
}

@article{hu2024metric3d,
  title={Metric3d v2: A versatile monocular geometric foundation model for zero-shot metric depth and surface normal estimation},
  author={Hu, Mu and Yin, Wei and Zhang, Chi and Cai, Zhipeng and Long, Xiaoxiao and Chen, Hao and Wang, Kaixuan and Yu, Gang and Shen, Chunhua and Shen, Shaojie},
  journal={IEEE Transactions on Pattern Analysis and Machine Intelligence},
  year={2024},
  publisher={IEEE}
}

@inproceedings{
bochkovskii2024depth,
title={Depth Pro: Sharp Monocular Metric Depth in Less Than a Second},
author={Alexey Bochkovskiy and Ama{\"e}l Delaunoy and Hugo Germain and Marcel Santos and Yichao Zhou and Stephan Richter and Vladlen Koltun},
booktitle={International Conference on Learning Representations (ICLR)},
year={2025},
}

@inproceedings{wang2025moge,
  title={Moge: Unlocking accurate monocular geometry estimation for open-domain images with optimal training supervision},
  author={Wang, Ruicheng and Xu, Sicheng and Dai, Cassie and Xiang, Jianfeng and Deng, Yu and Tong, Xin and Yang, Jiaolong},
  booktitle={Proceedings of the IEEE/CVF Conference on Computer Vision and Pattern Recognition (CVPR)},
  pages={5261--5271},
  year={2025}
}

@inproceedings{xiang2025structured,
  title={Structured 3d latents for scalable and versatile 3d generation},
  author={Xiang, Jianfeng and Lv, Zelong and Xu, Sicheng and Deng, Yu and Wang, Ruicheng and Zhang, Bowen and Chen, Dong and Tong, Xin and Yang, Jiaolong},
  booktitle={Proceedings of the IEEE/CVF Conference on Computer Vision and Pattern Recognition (CVPR)},
  pages={21469--21480},
  year={2025}
}

@inproceedings{azinovic2022neural,
  title={Neural rgb-d surface reconstruction},
  author={Azinovi{\'c}, Dejan and Martin-Brualla, Ricardo and Goldman, Dan B and Nie{\ss}ner, Matthias and Thies, Justus},
  booktitle={Proceedings of the IEEE/CVF Conference on Computer Vision and Pattern Recognition (CVPR)},
  pages={6290--6301},
  year={2022}
}

@inproceedings{boss2025sf3d,
  title={Sf3d: Stable fast 3d mesh reconstruction with uv-unwrapping and illumination disentanglement},
  author={Boss, Mark and Huang, Zixuan and Vasishta, Aaryaman and Jampani, Varun},
  booktitle={Proceedings of the IEEE/CVF Conference on Computer Vision and Pattern Recognition (CVPR)},
  pages={16240--16250},
  year={2025}
}

@inproceedings{huang2025spar3d,
  title={Spar3d: Stable point-aware reconstruction of 3d objects from single images},
  author={Huang, Zixuan and Boss, Mark and Vasishta, Aaryaman and Rehg, James M and Jampani, Varun},
  booktitle={Proceedings of the IEEE/CVF Conference on Computer Vision and Pattern Recognition (CVPR)},
  pages={16860--16870},
  year={2025}
}

@inproceedings{shotton2013scene,
  title={Scene coordinate regression forests for camera relocalization in RGB-D images},
  author={Shotton, Jamie and Glocker, Ben and Zach, Christopher and Izadi, Shahram and Criminisi, Antonio and Fitzgibbon, Andrew},
  booktitle={Proceedings of the IEEE/CVF Conference on Computer Vision and Pattern Recognition (CVPR)},
  pages={2930--2937},
  year={2013}
}

@inproceedings{
dehghan2021arkitscenes,
title={{ARK}itScenes - A Diverse Real-World Dataset for 3D Indoor Scene Understanding Using Mobile {RGB}-D Data},
author={Gilad Baruch and Zhuoyuan Chen and Afshin Dehghan and Tal Dimry and Yuri Feigin and Peter Fu and Thomas Gebauer and Brandon Joffe and Daniel Kurz and Arik Schwartz and Elad Shulman},
booktitle={Thirty-fifth Conference on Neural Information Processing Systems Datasets and Benchmarks Track},
year={2021},
}

@InProceedings{Sucar_dpm,
    author    = {Sucar, Edgar and Lai, Zihang and Insafutdinov, Eldar and Vedaldi, Andrea},
    title     = {Dynamic Point Maps: A Versatile Representation for Dynamic 3D Reconstruction},
    booktitle = {International Conference on Computer Vision (ICCV)},
    month     = {October},
    year      = {2025},
    pages     = {7295-7305}
}

@inproceedings{st4rtrack2025,
    title={St4RTrack: Simultaneous 4D Reconstruction and Tracking in the World},
    author={Feng, Haiwen and Zhang, Junyi and Wang, Qianqian and Ye, Yufei and Yu, Pengcheng and Black, Michael J. and Darrell, Trevor and Kanazawa, Angjoo},
    booktitle={International Conference on Computer Vision (ICCV)},
    year={2025}
}

@String(CVPR= {IEEE Conf. Comput. Vis. Pattern Recog.})

@String(ICCV= {Int. Conf. Comput. Vis.})

@String(ECCV= {Eur. Conf. Comput. Vis.})

@String(TOG= {ACM Trans. Graph.})

@String(ICLR = {Int. Conf. Learn. Represent.})

@String(CVPR  = {CVPR})

@String(ICCV  = {ICCV})

@String(ECCV  = {ECCV})

@String(TOG   = {ACM TOG})

@String(ICLR  = {ICLR})

@inproceedings{Silberman:ECCV12,
  author    = {Nathan Silberman, Derek Hoiem, Pushmeet Kohli and Rob Fergus},
  title     = {Indoor Segmentation and Support Inference from RGBD Images},
  booktitle={European Conference on Computer Vision (ECCV)},
  year      = {2012}
}

@article{chang2015shapenet,
  title={Shapenet: An information-rich 3d model repository},
  author={Chang, Angel X and Funkhouser, Thomas and Guibas, Leonidas and Hanrahan, Pat and Huang, Qixing and Li, Zimo and Savarese, Silvio and Savva, Manolis and Song, Shuran and Su, Hao and others},
  journal={arXiv preprint arXiv:1512.03012},
  year={2015}
}

@inproceedings{eigen2015predicting,
  title={Predicting depth, surface normals and semantic labels with a common multi-scale convolutional architecture},
  author={Eigen, David and Fergus, Rob},
  booktitle={International Conference on Computer Vision (ICCV)},
  pages={2650--2658},
  year={2015}
}

@inproceedings{choy20163d,
  title={3d-r2n2: A unified approach for single and multi-view 3d object reconstruction},
  author={Choy, Christopher B and Xu, Danfei and Gwak, JunYoung and Chen, Kevin and Savarese, Silvio},
  booktitle={European conference on computer vision (ECCV)},
  pages={628--644},
  year={2016},
  organization={Springer}
}

@inproceedings{fan2017point,
  title={A point set generation network for 3d object reconstruction from a single image},
  author={Fan, Haoqiang and Su, Hao and Guibas, Leonidas J},
  booktitle={Proceedings of the IEEE Conference on Computer Vision and Pattern Recognition (CVPR)},
  pages={605--613},
  year={2017}
}

@article{vaswani2017attention,
  title={Attention is all you need},
  author={Vaswani, Ashish and Shazeer, Noam and Parmar, Niki and Uszkoreit, Jakob and Jones, Llion and Gomez, Aidan N and Kaiser, {\L}ukasz and Polosukhin, Illia},
  journal={Advances in Neural Information Processing Systems (NeurIPS)},
  volume={30},
  year={2017}
}

@inproceedings{wang2018pixel2mesh,
  title={Pixel2mesh: Generating 3d mesh models from single rgb images},
  author={Wang, Nanyang and Zhang, Yinda and Li, Zhuwen and Fu, Yanwei and Liu, Wei and Jiang, Yu-Gang},
  booktitle={European Conference on Computer Vision (ECCV)},
  pages={52--67},
  year={2018}
}

@inproceedings{mildenhall2020nerf,
  title={Nerf: Representing scenes as neural radiance fields for view synthesis},
  author={Mildenhall, B and Srinivasan, PP and Tancik, M and Barron, JT and Ramamoorthi, R and Ng, R},
  booktitle={European conference on computer vision (ECCV)},
  year={2020}
}

@inproceedings{rombach2022high,
  title={High-resolution image synthesis with latent diffusion models},
  author={Rombach, Robin and Blattmann, Andreas and Lorenz, Dominik and Esser, Patrick and Ommer, Bj{\"o}rn},
  booktitle={Proceedings of the IEEE/CVF Conference on Computer Vision and Pattern Recognition (CVPR)},
  pages={10684--10695},
  year={2022}
}

@article{vahdat2022lion,
  title={Lion: Latent point diffusion models for 3d shape generation},
  author={Vahdat, Arash and Williams, Francis and Gojcic, Zan and Litany, Or and Fidler, Sanja and Kreis, Karsten and others},
  journal={Advances in Neural Information Processing Systems (NeurIPS)},
  volume={35},
  pages={10021--10039},
  year={2022}
}

@inproceedings{lipman2023flow,
  title={Flow Matching for Generative Modeling},
  author={Yaron Lipman and Ricky T. Q. Chen and Heli Ben-Hamu and Maximilian Nickel and Matthew Le},
  booktitle={The Eleventh International Conference on Learning Representations (ICLR)},
  year={2023},
}

@article{zhang20233dshape2vecset,
  title={3dshape2vecset: A 3d shape representation for neural fields and generative diffusion models},
  author={Zhang, Biao and Tang, Jiapeng and Niessner, Matthias and Wonka, Peter},
  journal={ACM Transactions On Graphics (TOG)},
  volume={42},
  number={4},
  pages={1--16},
  year={2023},
  publisher={ACM New York, NY, USA}
}

@inproceedings{ye2024consistent,
  title={Consistent-1-to-3: Consistent image to 3d view synthesis via geometry-aware diffusion models},
  author={Ye, Jianglong and Wang, Peng and Li, Kejie and Shi, Yichun and Wang, Heng},
  booktitle={2024 International Conference on 3D Vision (3DV)},
  pages={664--674},
  year={2024},
  organization={IEEE}
}

@inproceedings{deitke2023objaverse,
  title={Objaverse: A universe of annotated 3d objects},
  author={Deitke, Matt and Schwenk, Dustin and Salvador, Jordi and Weihs, Luca and Michel, Oscar and VanderBilt, Eli and Schmidt, Ludwig and Ehsani, Kiana and Kembhavi, Aniruddha and Farhadi, Ali},
  booktitle={Proceedings of the IEEE/CVF Conference on Computer Vision and Pattern Recognition (CVPR)},
  pages={13142--13153},
  year={2023}
}

@inproceedings{liu2023zero,
  title={Zero-1-to-3: Zero-shot one image to 3d object},
  author={Liu, Ruoshi and Wu, Rundi and Van Hoorick, Basile and Tokmakov, Pavel and Zakharov, Sergey and Vondrick, Carl},
  booktitle={International Conference on Computer Vision (ICCV)},
  pages={9298--9309},
  year={2023}
}

@article{kerbl20233d,
  title={3D Gaussian Splatting for Real-Time Radiance Field Rendering.},
  author={Kerbl, Bernhard and Kopanas, Georgios and Leimk{\"u}hler, Thomas and Drettakis, George},
  journal={ACM Trans. Graph.},
  volume={42},
  number={4},
  pages={139--1},
  year={2023}
}

@article{zhao2023michelangelo,
  title={Michelangelo: Conditional 3d shape generation based on shape-image-text aligned latent representation},
  author={Zhao, Zibo and Liu, Wen and Chen, Xin and Zeng, Xianfang and Wang, Rui and Cheng, Pei and Fu, Bin and Chen, Tao and Yu, Gang and Gao, Shenghua},
  journal={Advances in neural information processing systems (NeurIPS)},
  volume={36},
  pages={73969--73982},
  year={2023}
}

@article{liu2023one,
  title={One-2-3-45: Any single image to 3d mesh in 45 seconds without per-shape optimization},
  author={Liu, Minghua and Xu, Chao and Jin, Haian and Chen, Linghao and Varma T, Mukund and Xu, Zexiang and Su, Hao},
  journal={Advances in Neural Information Processing Systems (NeurIPS)},
  volume={36},
  pages={22226--22246},
  year={2023}
}

@inproceedings{hong2023lrm,
  title={Lrm: Large reconstruction model for single image to 3d},
  author={Hong, Yicong and Zhang, Kai and Gu, Jiuxiang and Bi, Sai and Zhou, Yang and Liu, Difan and Liu, Feng and Sunkavalli, Kalyan and Bui, Trung and Tan, Hao},
  booktitle={International Conference on Learning Representations (ICLR)},
  year={2024}
}

@inproceedings{
shi2024mvdream,
title={{MVD}ream: Multi-view Diffusion for 3D Generation},
author={Yichun Shi and Peng Wang and Jianglong Ye and Long Mai and Kejie Li and Xiao Yang},
booktitle={The Twelfth International Conference on Learning Representations (ICLR)},
year={2024},
url={https://openreview.net/forum?id=FUgrjq2pbB}
}

@inproceedings{li2024instant3d,
  author    = {Jiahao Li and Hao Tan and Kai Zhang and Zexiang Xu and Fujun
               Luan and Yinghao Xu and Yicong Hong and Kalyan Sunkavalli and
               Greg Shakhnarovich and Sai Bi},
  booktitle = {The Twelfth International Conference on Learning Representations (ICLR)},
  title     = {{Instant3D}: Fast Text-to-{3D} with Sparse-View Generation
               and Large Reconstruction Model},
  year      = {2024}
}

@inproceedings{wu2024reconfusion,
  title={Reconfusion: 3d reconstruction with diffusion priors},
  author={Wu, Rundi and Mildenhall, Ben and Henzler, Philipp and Park, Keunhong and Gao, Ruiqi and Watson, Daniel and Srinivasan, Pratul P and Verbin, Dor and Barron, Jonathan T and Poole, Ben and others},
  booktitle={Proceedings of the IEEE/CVF Conference on Computer Vision and Pattern Recognition (CVPR)},
  pages={21551--21561},
  year={2024}
}

@inproceedings{long2024wonder3d,
  title={Wonder3d: Single image to 3d using cross-domain diffusion},
  author={Long, Xiaoxiao and Guo, Yuan-Chen and Lin, Cheng and Liu, Yuan and Dou, Zhiyang and Liu, Lingjie and Ma, Yuexin and Zhang, Song-Hai and Habermann, Marc and Theobalt, Christian and others},
  booktitle={Proceedings of the IEEE/CVF Conference on Computer Vision and Pattern Recognition (CVPR)},
  pages={9970--9980},
  year={2024}
}

@inproceedings{sargent2024zeronvs,
  title={Zeronvs: Zero-shot 360-degree view synthesis from a single image},
  author={Sargent, Kyle and Li, Zizhang and Shah, Tanmay and Herrmann, Charles and Yu, Hong-Xing and Zhang, Yunzhi and Chan, Eric Ryan and Lagun, Dmitry and Fei-Fei, Li and Sun, Deqing and others},
  booktitle={Proceedings of the IEEE/CVF Conference on Computer Vision and Pattern Recognition (CVPR)},
  pages={9420--9429},
  year={2024}
}

@inproceedings{wang2024motionctrl,
  title={Motionctrl: A unified and flexible motion controller for video generation},
  author={Wang, Zhouxia and Yuan, Ziyang and Wang, Xintao and Li, Yaowei and Chen, Tianshui and Xia, Menghan and Luo, Ping and Shan, Ying},
  booktitle={ACM SIGGRAPH 2024 Conference Papers},
  pages={1--11},
  year={2024}
}

@inproceedings{zhang2024gs,
  title={Gs-lrm: Large reconstruction model for 3d gaussian splatting},
  author={Zhang, Kai and Bi, Sai and Tan, Hao and Xiangli, Yuanbo and Zhao, Nanxuan and Sunkavalli, Kalyan and Xu, Zexiang},
  booktitle={European Conference on Computer Vision (ECCV)},
  pages={1--19},
  year={2024},
  organization={Springer}
}

@inproceedings{han2024vfusion3d,
  title={Vfusion3d: Learning scalable 3d generative models from video diffusion models},
  author={Han, Junlin and Kokkinos, Filippos and Torr, Philip},
  booktitle={European Conference on Computer Vision (ECCV)},
  pages={333--350},
  year={2024},
  organization={Springer}
}

@inproceedings{tang2025lgm,
  title={Lgm: Large multi-view gaussian model for high-resolution 3d content creation},
  author={Tang, Jiaxiang and Chen, Zhaoxi and Chen, Xiaokang and Wang, Tengfei and Zeng, Gang and Liu, Ziwei},
  booktitle={European Conference on Computer Vision (ECCV)},
  pages={1--18},
  year={2025},
  organization={Springer}
}

@inproceedings{voleti2025sv3d,
  title={Sv3d: Novel multi-view synthesis and 3d generation from a single image using latent video diffusion},
  author={Voleti, Vikram and Yao, Chun-Han and Boss, Mark and Letts, Adam and Pankratz, David and Tochilkin, Dmitry and Laforte, Christian and Rombach, Robin and Jampani, Varun},
  booktitle={European Conference on Computer Vision (ECCV)},
  pages={439--457},
  year={2024},
  organization={Springer}
}

@inproceedings{wang2024dust3r,
  title={Dust3r: Geometric 3d vision made easy},
  author={Wang, Shuzhe and Leroy, Vincent and Cabon, Yohann and Chidlovskii, Boris and Revaud, Jerome},
  booktitle={Proceedings of the IEEE/CVF Conference on Computer Vision and Pattern Recognition (CVPR)},
  pages={20697--20709},
  year={2024}
}

@inproceedings{ren2024xcube,
  title={Xcube: Large-scale 3d generative modeling using sparse voxel hierarchies},
  author={Ren, Xuanchi and Huang, Jiahui and Zeng, Xiaohui and Museth, Ken and Fidler, Sanja and Williams, Francis},
  booktitle={Proceedings of the IEEE/CVF Conference on Computer Vision and Pattern Recognition (CVPR)},
  pages={4209--4219},
  year={2024}
}

@inproceedings{leroy2024grounding,
  title={Grounding image matching in 3d with mast3r},
  author={Leroy, Vincent and Cabon, Yohann and Revaud, J{\'e}r{\^o}me},
  booktitle={European Conference on Computer Vision (ECCV)},
  pages={71--91},
  year={2024},
  organization={Springer}
}

@article{gao2024cat3d,
  title={Cat3d: Create anything in 3d with multi-view diffusion models},
  author={Gao, Ruiqi and Holynski, Aleksander and Henzler, Philipp and Brussee, Arthur and Martin-Brualla, Ricardo and Srinivasan, Pratul and Barron, Jonathan T and Poole, Ben},
  journal={Advances in Neural Information Processing Systems (NeurIPS)},
  year={2024}
}

@article{wallingford2024image,
  title={From an Image to a Scene: Learning to Imagine the World from a Million 360° Videos},
  author={Wallingford, Matthew and Bhattad, Anand and Kusupati, Aditya and Ramanujan, Vivek and Deitke, Matt and Kembhavi, Aniruddha and Mottaghi, Roozbeh and Ma, Wei-Chiu and Farhadi, Ali},
  journal={Advances in Neural Information Processing Systems (NeurIPS)},
  volume={37},
  pages={17743--17760},
  year={2024}
}

@article{zhang2024clay,
  title={CLAY: A Controllable Large-scale Generative Model for Creating High-quality 3D Assets},
  author={Zhang, Longwen and Wang, Ziyu and Zhang, Qixuan and Qiu, Qiwei and Pang, Anqi and Jiang, Haoran and Yang, Wei and Xu, Lan and Yu, Jingyi},
  journal={ACM Transactions on Graphics (TOG)},
  volume={43},
  number={4},
  pages={1--20},
  year={2024},
  publisher={ACM New York, NY, USA}
}

@inproceedings{zhang2024world,
  title={World-consistent Video Diffusion with Explicit 3D Modeling},
  author={Zhang, Qihang and Zhai, Shuangfei and Bautista, Miguel Angel and Miao, Kevin and Toshev, Alexander and Susskind, Joshua and Gu, Jiatao},
  booktitle={Proceedings of the IEEE/CVF Conference on Computer Vision and Pattern Recognition (CVPR)},
  year={2025}
}

@InProceedings{Yang_2025_Fast3R,
    title={Fast3R: Towards 3D Reconstruction of 1000+ Images in One Forward Pass},
    author={Jianing Yang and Alexander Sax and Kevin J. Liang and Mikael Henaff and Hao Tang and Ang Cao and Joyce Chai and Franziska Meier and Matt Feiszli},
    booktitle={Proceedings of the IEEE/CVF Conference on Computer Vision and Pattern Recognition (CVPR)},
    month={June},
    year={2025},
}

@InProceedings{tang2025mv,
  title={MV-DUSt3R+: Single-Stage Scene Reconstruction from Sparse Views In 2 Seconds},
  author={Tang, Zhenggang and Fan, Yuchen and Wang, Dilin and Xu, Hongyu and Ranjan, Rakesh and Schwing, Alexander and Yan, Zhicheng},
  booktitle={Proceedings of the IEEE/CVF Conference on Computer Vision and Pattern Recognition (CVPR)},
  month={June},
  year={2025},
}

@inproceedings{yu2025wonderworld,
    title={WonderWorld: Interactive 3D Scene Generation from a Single Image},
    author={Hong-Xing Yu and Haoyi Duan and Charles Herrmann and William T. Freeman and Jiajun Wu},
    booktitle={Proceedings of the IEEE/CVF Conference on Computer Vision and Pattern Recognition (CVPR)},
    month={June},
    year={2025},
}

@InProceedings{wang2025continuous,
  title={Continuous 3D Perception Model with Persistent State},
  author={Wang, Qianqian and Zhang, Yifei and Holynski, Aleksander and Efros, Alexei A and Kanazawa, Angjoo},
  booktitle={Proceedings of the IEEE/CVF Conference on Computer Vision and Pattern Recognition (CVPR)},
  month={June},
  year={2025},
}

@inproceedings{zhang2025FLARE,
    title =
    {FLARE: Feed-forward Geometry, Appearance and Camera Estimation from Uncalibrated Sparse Views},
    author =
    {Shangzhan Zhang and Jianyuan Wang and Yinghao Xu and Nan Xue and Christian Rupprecht and Xiaowei Zhou and Yujun Shen and Gordon Wetzstein},
    booktitle={Proceedings of the IEEE/CVF Conference on Computer Vision and Pattern Recognition (CVPR)},
  month={June},
  year={2025},
}

@inproceedings{xiang2024structured,
    title   = {Structured 3D Latents for Scalable and Versatile 3D Generation},
    author  = {Xiang, Jianfeng and Lv, Zelong and Xu, Sicheng and Deng, Yu and Wang, Ruicheng and Zhang, Bowen and Chen, Dong and Tong, Xin and Yang, Jiaolong},
    booktitle={Proceedings of the IEEE/CVF Conference on Computer Vision and Pattern Recognition (CVPR)},
    month={June},
    year={2025}
}

@inproceedings{wang2025vggt,
  title={Vggt: Visual geometry grounded transformer},
  author={Wang, Jianyuan and Chen, Minghao and Karaev, Nikita and Vedaldi, Andrea and Rupprecht, Christian and Novotny, David},
  booktitle={Proceedings of the Computer Vision and Pattern Recognition Conference (CVPR)},
  pages={5294--5306},
  year={2025}
}

@inproceedings{szymanowicz2025bolt3d,
  title={Bolt3D: Generating 3D Scenes in Seconds},
  author={Szymanowicz, Stanislaw and Zhang, Jason Y and Srinivasan, Pratul and Gao, Ruiqi and Brussee, Arthur and Holynski, Aleksander and Martin-Brualla, Ricardo and Barron, Jonathan T and Henzler, Philipp},
  booktitle   = {International Conference on Computer Vision (ICCV)},
  month     = {October},
  year      = {2025}
}

@inproceedings{jiang2025rayzer,
  title={RayZer: A Self-supervised Large View Synthesis Model},
  author={Jiang, Hanwen and Tan, Hao and Wang, Peng and Jin, Haian and Zhao, Yue and Bi, Sai and Zhang, Kai and Luan, Fujun and Sunkavalli, Kalyan and Huang, Qixing and others},
  booktitle   = {International Conference on Computer Vision (ICCV)},
  month     = {October},
  year      = {2025}
}

@article{blattmann2023stable,
  title={Stable video diffusion: Scaling latent video diffusion models to large datasets},
  author={Blattmann, Andreas and Dockhorn, Tim and Kulal, Sumith and Mendelevitch, Daniel and Kilian, Maciej and Lorenz, Dominik and Levi, Yam and English, Zion and Voleti, Vikram and Letts, Adam and others},
  journal={arXiv preprint arXiv:2311.15127},
  year={2023}
}

@article{TripoSR2024,
  title={TripoSR: Fast 3D Object Reconstruction from a Single Image},
  author={Tochilkin, Dmitry and Pankratz, David and Liu, Zexiang and Huang, Zixuan and and Letts, Adam and Li, Yangguang and Liang, Ding and Laforte, Christian and Jampani, Varun and Cao, Yan-Pei},
  journal={arXiv preprint arXiv:2403.02151},
  year={2024}
}

@article{yu2024viewcrafter,
  title={ViewCrafter: Taming Video Diffusion Models for High-fidelity Novel View Synthesis},
  author={Yu, Wangbo and Xing, Jinbo and Yuan, Li and Hu, Wenbo and Li, Xiaoyu and Huang, Zhipeng and Gao, Xiangjun and Wong, Tien-Tsin and Shan, Ying and Tian, Yonghong},
  journal={IEEE Transactions on Pattern Analysis \& Machine Intelligence},
  number={01},
  pages={1--18},
  year={2025},
  publisher={IEEE Computer Society}
}

@misc{liu2024reconx,
        title={ReconX: Reconstruct Any Scene from Sparse Views with Video Diffusion Model}, 
        author={Fangfu Liu and Wenqiang Sun and Hanyang Wang and Yikai Wang and Haowen Sun and Junliang Ye and Jun Zhang and Yueqi Duan},
        year={2024},
        eprint={2408.16767},
        archivePrefix={arXiv},
        primaryClass={cs.CV},
        url={https://arxiv.org/abs/2408.16767},
      }

@misc{cabon2020vkitti2,
  title={Virtual KITTI 2},
  author={Cabon, Yohann and Murray, Naila and Humenberger, Martin},
  year={2020},
  eprint={2001.10773},
  archivePrefix={arXiv},
  primaryClass={cs.CV}
}

@misc{hunyuan3d22025tencent,
    title={Hunyuan3D 2.0: Scaling Diffusion Models for High Resolution Textured 3D Assets Generation},
    author={Tencent Hunyuan3D Team},
    year={2025},
    eprint={2501.12202},
    archivePrefix={arXiv},
    primaryClass={cs.CV}
}

@misc{yang2024hunyuan3d,
    title={Hunyuan3D 1.0: A Unified Framework for Text-to-3D and Image-to-3D Generation},
    author={Tencent Hunyuan3D Team},
    year={2024},
    eprint={2411.02293},
    archivePrefix={arXiv},
    primaryClass={cs.CV}
}

@article{li2025triposg,
  title={TripoSG: High-Fidelity 3D Shape Synthesis using Large-Scale Rectified Flow Models},
  author={Li, Yangguang and Zou, Zi-Xin and Liu, Zexiang and Wang, Dehu and Liang, Yuan and Yu, Zhipeng and Liu, Xingchao and Guo, Yuan-Chen and Liang, Ding and Ouyang, Wanli and others},
  journal={arXiv preprint arXiv:2502.06608},
  year={2025}
}

@inproceedings{zhou2025stable,
  title={Stable virtual camera: Generative view synthesis with diffusion models},
  author={Zhou, Jensen and Gao, Hang and Voleti, Vikram and Vasishta, Aaryaman and Yao, Chun-Han and Boss, Mark and Torr, Philip and Rupprecht, Christian and Jampani, Varun},
  booktitle={Proceedings of the IEEE/CVF International Conference on Computer Vision (CVPR)},
  pages={12405--12414},
  year={2025}
}

@inproceedings{wu2025amodal3r,
      title={Amodal3R: Amodal 3D Reconstruction from Occluded 2D Images}, 
      author={Tianhao Wu and Chuanxia Zheng and Frank Guan and Andrea Vedaldi and Tat-Jen Cham},
      booktitle={International Conference on Computer Vision (ICCV)},
      year={2025}
}

@inproceedings{chen2024mvsplat360,
  title={MVSplat360: Feed-Forward 360 Scene Synthesis from Sparse Views},
  author={Chen, Yuedong and Zheng, Chuanxia and Xu, Haofei and Zhuang, Bohan and Vedaldi, Andrea and Cham, Tat-Jen and Cai, Jianfei},
  booktitle={Advances in Neural Information Processing Systems (NeurIPS)},
  year={2024}
}

@inproceedings{zheng2024free3d,
  title={Free3d: Consistent novel view synthesis without 3d representation},
  author={Zheng, Chuanxia and Vedaldi, Andrea},
  booktitle={Proceedings of the IEEE/CVF Conference on Computer Vision and Pattern Recognition (CVPR)},
  pages={9720--9731},
  year={2024}
}

\newpage
\appendix
\section{Appendix}

\subsection{More implementation details.}
\paragraph{Model architectures.} For the 3D point autoencoder (Stage 1), we follow the point encoder design from TripoSG~\cite{li2025triposg}, which consists of one cross-attention layer and eight self-attention layers. The initial point queries are obtained by farthest point sampling from the input point cloud, while the learnable queries are randomly initialized tokens. We use 768 tokens with dimension 128 for the our model. For the flow-matching decoder, we use a joint block with two cross-attention layers and one self-attention layer. The goal is to enable self-attention–like information exchange among queries while keeping computation manageable. Concretely, each block (i) aggregates information from noisy query points into the scene tokens via cross-attention, (ii) performs self-attention on the scene tokens (small $M$) to mix global context efficiently, and (iii) projects the updated scene tokens back to the queries with a second cross-attention.

For the image-to-latent transformer in Stage 2, we follow the architecture of VGGT~\citep{wang2025vggt}, which alternates between local (frame-level) and global attention. Due to computational constraints, we adopt a 16-layer variant instead of the full 24-layer VGGT, initializing from its pretrained weights. We also reuse VGGT’s image tokenizers with frozen weights to obtain image tokens. The initial 3D scene tokens are treated as a \emph{3D frame} and share the same local attention mechanism with the image tokens. For the 3D scene token, we copy the camera token from the first view to enable reconstruction in the camera coordinate of the first view. 

\paragraph{Training.}
We train our model in two stages. In Stage 1, we aggregate per-view point clouds into a single input cloud and apply farthest point sampling on a randomly selected subset to supervise the flow-matching decoder. Farthest point sampling ensures that the target point cloud is distributed more evenly, reducing the influence of overlapping points in visible regions. Stage 1 is trained for 50 epochs. In Stage 2, we reuse the flow-matching decoder from Stage 1 and train it together with our image encoder, initialized with pretrained VGGT weights. The same flow-matching loss is used in both stages. For object and scene completion, target point clouds are sampled from complete reconstructions. To demonstrate compatibility with pixel-aligned formats, we also train a variant using RGB-D input, where target point clouds are sampled from point maps back-projected from depth. Stage 2 is trained for another 50 epochs.

\rebuttal{Regarding computational cost, the Stage-1 point encoder is lightweight and requires no paired image–point cloud data, enabling efficient training on large-scale synthetic 3D datasets. In practice, Stage 1 takes about 40\% less training time than Stage 2, and inference remains single-stage, feed-forward, and efficient regardless of the two-stage setup. Overall, the two-stage design adds small overhead while substantially improving stability, data flexibility, and reconstruction quality.}

\paragraph{Evaluation.} For object- and scene-level completion, we follow \citet{li2025lari} and sample 10k points for the object task and 100k for the scene task. However, correspondence-based point cloud alignment is not applicable due to our non–pixel-aligned reconstruction. Instead, we optimize a 3D translation and a global (1D) scale relative to the ground-truth point cloud using Adam to improve alignment. We do not optimize rotation, as our reconstruction is expressed in the first-view coordinate frame.

\subsection{More ablation study.}

\begin{figure*}[h]
    \centering
    \includegraphics[width=\linewidth]{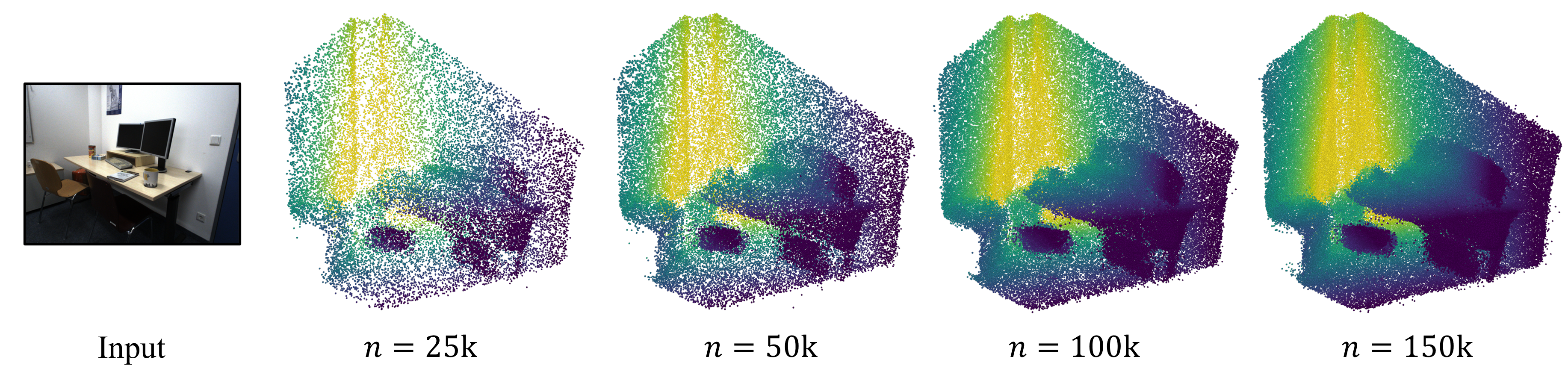}
    \caption{\textbf{Visualization of point cloud generation at different resolutions.}
    Our non–pixel-aligned formulation allows inference at arbitrary resolutions.}%
    \label{fig:query_nuber_vis}
\end{figure*}

\paragraph{Reconstruction at any resolution.}
Since NOVA3R models a point distribution rather than a per-pixel point map,
it naturally  supports resolution-agnostic generation by adjusting the number of noisy queries at inference.
\Cref{fig:query_nuber_vis} presents results with varying query counts for the flow-matching decoder,
demonstrating that our method consistently produces point clouds at different resolutions with reliable reconstruction quality.

\subsection{More visualizations.}
\begin{figure*}[tb!]
    \centering
    \includegraphics[width=\linewidth]{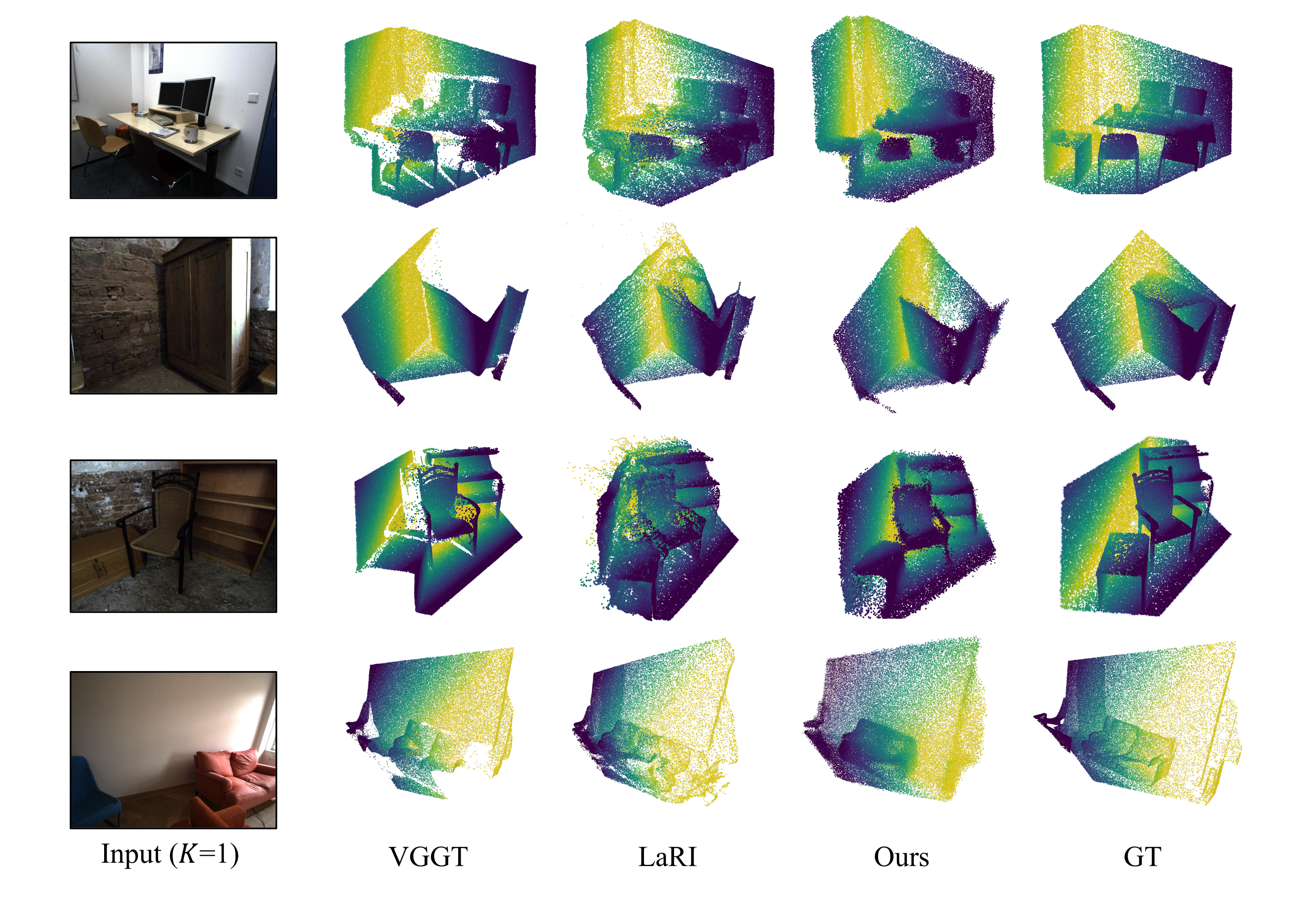}
    \caption{\textbf{Qualitative results for scene completion on SCRREAM~\cite{jung2024scrream}}. Our method shows better scene completion results compared to other baselines.}%
    \label{fig:scrream_n1_supp}
\end{figure*}

\begin{figure*}[tb!]
    \centering
    \includegraphics[width=\linewidth]{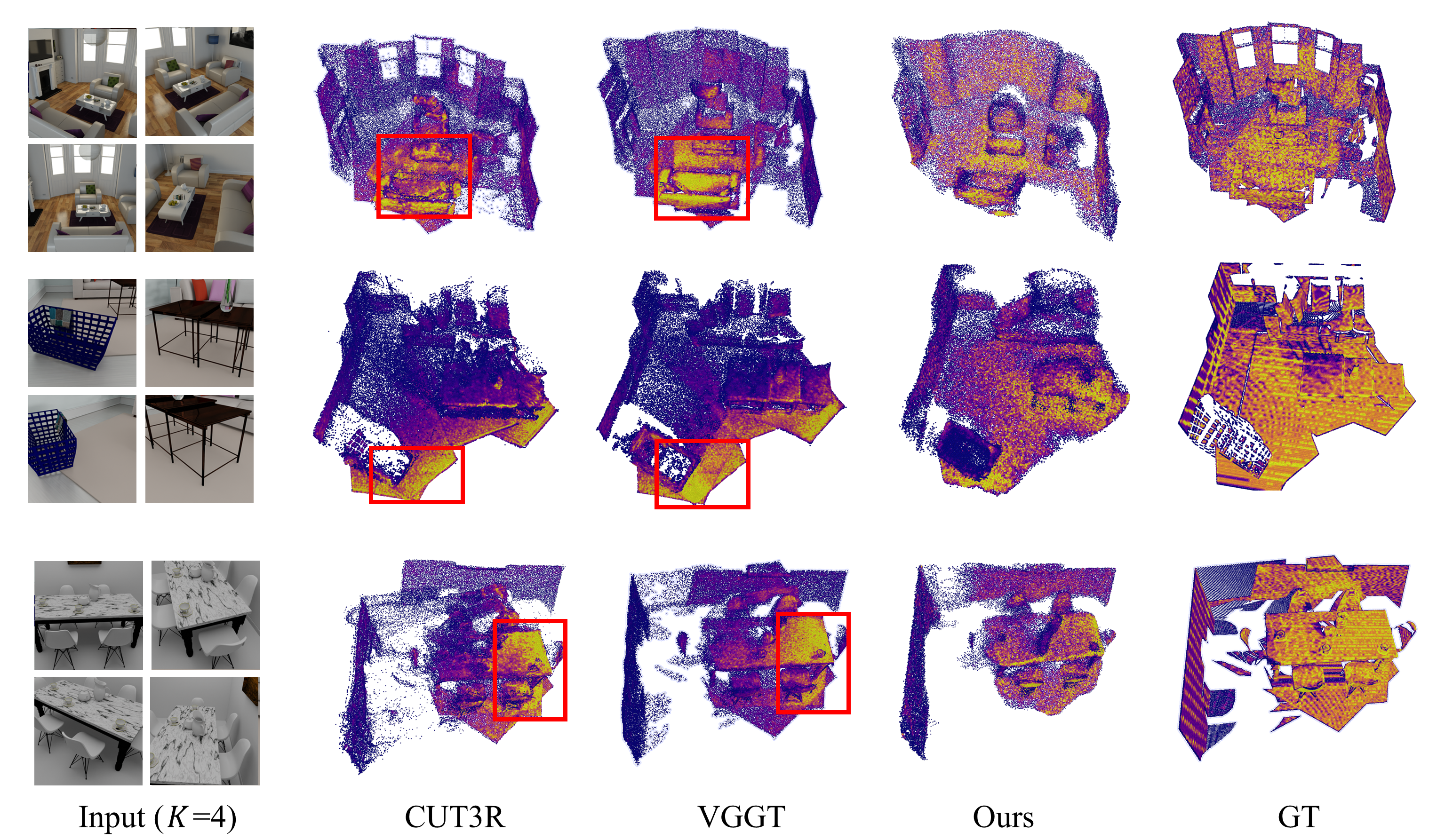}
    \caption{\textbf{Qualitative results for density evaluation on NRGBD ($K{=}4$)~\citep{azinovic2022neural}.}
    Yellow regions denote higher density, and purple regions denote lower density. Our method provides more evenly-distributed point cloud (colored by density).}%
    \label{fig:nrgbd_supp}
\end{figure*}

We show more visualization results for scene-level completion on SCRREAM~\citep{jung2024scrream} dataset, as shonw in~\Cref{fig:scrream_n1_supp}. We also include density evaluation on NRGBD~\citep{azinovic2022neural} in~\Cref{fig:nrgbd_supp}. While trained with $K=2$ views only, our method generalize to multiple image views $(K=4)$ and provides more evenly distributed point cloud .

\subsection{Reducing Uncertainty in Latent Diffusion–Based 3D Generation}
Our method is specifically designed to reduce the uncertainty typically observed in latent diffusion–based 3D generation approaches such as TRELLIS~\citep{xiang2024structured} and TripoSG~\citep{li2025triposg}. These methods perform generation in a high-dimensional latent space, which may lead to hallucinated geometry, shape deviations, and inconsistencies across viewpoints—particularly when multiple input images are involved. As a result, they struggle to maintain strong pixel-to-scene and cross-view alignment.

\begin{figure*}[tb!]
    \centering
    \includegraphics[width=\linewidth]{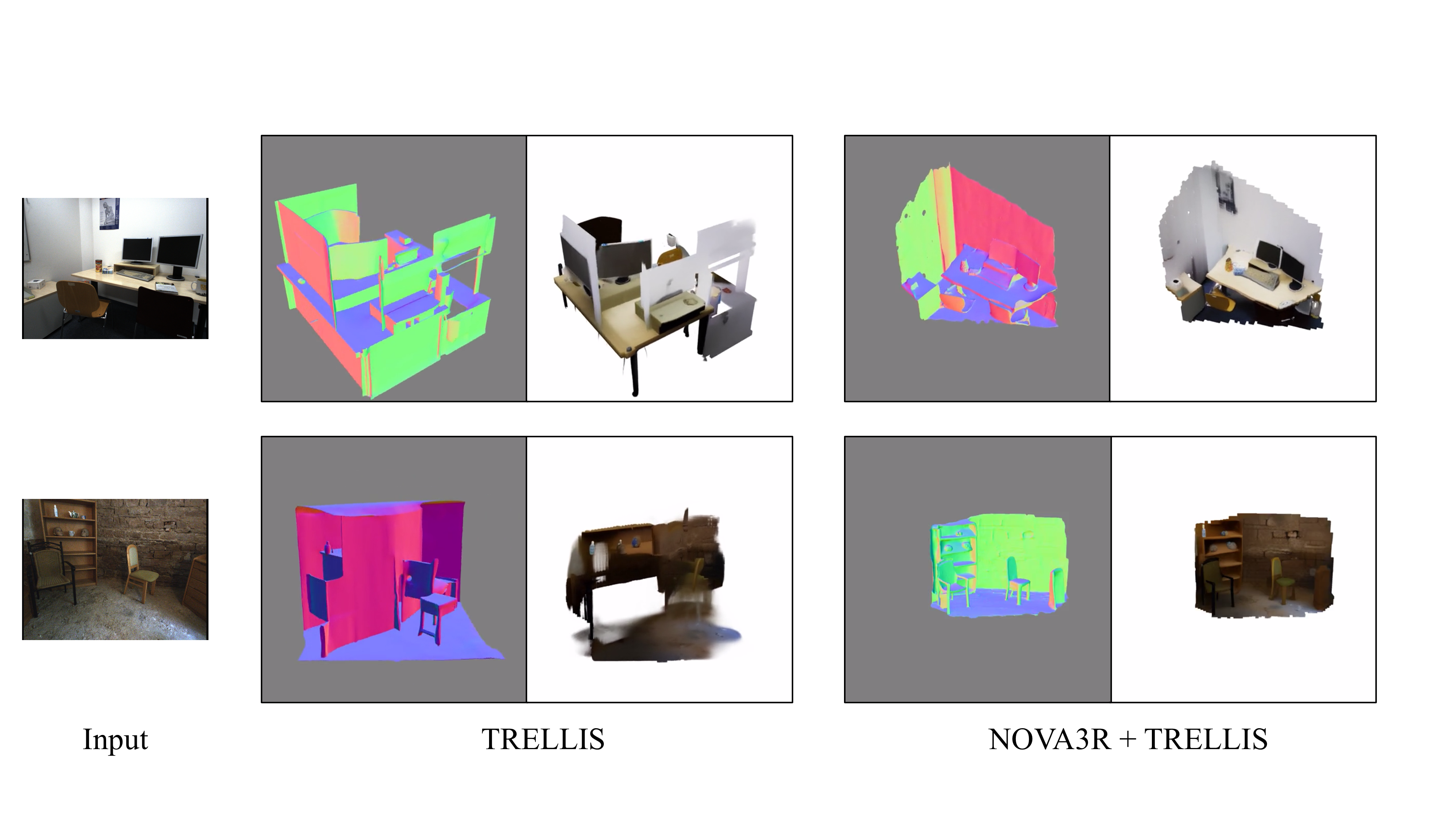}
    \caption{\rebuttal{\textbf{Extending 3D object generation model to real-world scene reconstruction.}
    The pretrained TRELLIS model struggles to generate geometrically faithful reconstructions for real-world scenes. In contrast, NOVA3R can provide active voxel priors for the TRELLIS stage-1 generation process, enabling its extension to real-world scene synthesis. }}%
    \label{fig:trellis}
\end{figure*}

In comparison, NOVA3R provides faithful reconstruction conditioned on the input images. Furthermore, NOVA3R can be integrated with the pretrained TRELLIS model to provide active voxel positions, effectively extending 3D object generation models to real-world scene synthesis without re-training (see~\Cref{fig:trellis}).
 
\subsection{Performance on Outdoor Scenes}

\begin{figure*}[tb!]
    \centering
    \includegraphics[width=\linewidth]{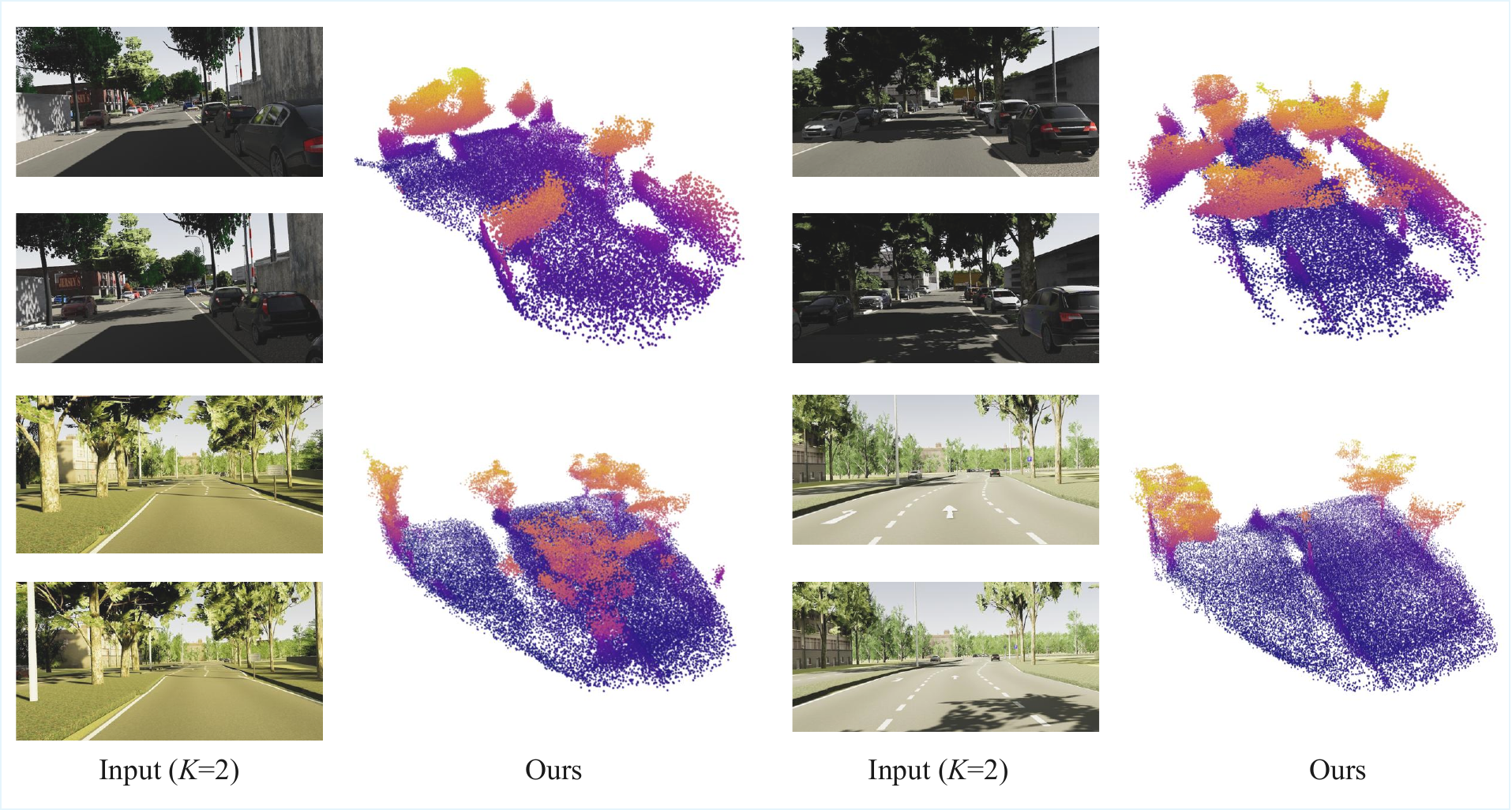}
    \caption{\rebuttal{\textbf{Qualitative results for outdoor scenes reconstruction on Virtual KITTI 2.}
    Our method is also applicable to outdoor scene reconstruction (colored by Y axis).}}%
    \label{fig:vkitti}
    \vspace{-15pt}
\end{figure*}

\rebuttal{To validate the robustness and generalization capability of our framework, we further evaluate NOVA3R using the outdoor dataset Virtual KITTI 2~\citep{cabon2020vkitti2}. We finetune our model on Virtual KITTI 2 to better adapt to large-scale outdoor environments.
To construct pseudo ground truth, for each input frame we collect neighboring frames within [-4,8] timesteps and additional views from $\pm 15^{\circ}$ and $\pm 30^{\circ}$ viewpoints. Using depth maps and camera parameters, we project them into per-frame point clouds, transform them to world coordinates, and retain only points within the target view’s frustum. As shown in~\Cref{fig:vkitti}, NOVA3R performs well on outdoor scenes, further demonstrating its ability to handle both indoor and outdoor scenarios.}

\subsection{Discussion}

\paragraph{Large-scale Scenes.} \rebuttal{Modeling large-scale scenes with many input images is a major computational bottleneck for existing learning-based 3D reconstruction methods, particularly for pixel-aligned approaches like VGGT, which must handle duplicated points across multiple views. In contrast, our point-wise decoding uses fewer tokens to represent the scene, making it inherently more scalable. However, the number of points needed varies across scenes of different scales, requiring adaptive point selection strategies, such as using sparse COLMAP point clouds as guidance. 
}

\paragraph{Dynamic Scenes.} 
\rebuttal{Our paradigm is inherently extensible to dynamic scenes, either by adding a branch to predict target time point maps~\citep{Sucar_dpm, st4rtrack2025}
or by extending the 3D latent autoencoder to a time-conditioned 4D latent representation. Such a representation can potentially model the entire 4D scene more efficiently by capturing both complete geometry and temporal evolution across the whole sequence, rather than relying on per-frame reconstruction. 
}

\end{document}